\documentclass[letterpaper]{article} 
\usepackage{aaai25}  
\usepackage{times}  
\usepackage{helvet}  
\usepackage{courier}  
\usepackage[hyphens]{url}  
\usepackage{graphicx} 
\usepackage{natbib}  
\usepackage{caption} 
\usepackage[table,dvipsnames]{xcolor}
\usepackage{custom_defs}
\usepackage{booktabs}       
\usepackage{tabularx}
\usepackage{multirow}
\usepackage{amsmath,bm}
\usepackage{makecell}
\usepackage{mwe}
\usepackage{nicefrac}
\usepackage{subcaption}

\definecolor{bestcol}{RGB}{254,196,79}
\newcommand{\best}[1]{{\cellcolor{bestcol}}\bfseries#1}
\definecolor{secondbestcol}{RGB}{255,247,188}
\newcommand{\secondbest}[1]{{\cellcolor{secondbestcol}}#1}

\usepackage{etoolbox,siunitx}
\robustify\bfseries
\usepackage[capitalize]{cleveref}

\nocopyright 

\title{High-Resolution Frame Interpolation with \ourCascade{} Cascaded Diffusion}
\author{
    Junhwa Hur\equalcontrib,\: 
    Charles Herrmann\equalcontrib,\:
    Saurabh Saxena,\:
    Janne Kontkanen,\:
    Wei-Sheng Lai,\:
    Yichang Shih,\:
    Michael Rubinstein,\:
    David J. Fleet\thanks{DF is also affiliated with the University of Toronto and the Vector Institute.},\:
    Deqing Sun
}
\affiliations{
    Google
}

\begin{document}

\maketitle

\begin{abstract}
    
Despite the recent progress, existing frame interpolation methods still struggle with processing extremely high resolution input and handling challenging cases such as repetitive textures, thin objects, and large motion.
To address these issues, we introduce a \emph{\ourcascade{}} cascaded pixel diffusion model for high resolution frame interpolation, \ourmethod{}, that excels in these scenarios while achieving competitive performance on standard benchmarks. 
Cascades, which generate a series of images from low to high resolution, can help significantly with large or complex motion that require both global context for a coarse solution and detailed context for high resolution output.
However, contrary to prior work on cascaded diffusion models which perform diffusion on increasingly large resolutions, we use a single model that always performs diffusion at the same resolution and upsamples by processing patches of the inputs and the prior solution.
At inference time, this drastically reduces memory usage and allows a single model, solving both frame interpolation (base model’s task) and spatial up-sampling, saving training cost as well.
\ourmethod{} excels at high-resolution images and complex repeated textures that require global context, achieving comparable or state-of-the-art performance on various benchmarks (Vimeo, Xiph, X-Test, and SEPE-8K).
We further introduce a new dataset, \ourdataset{}, that focuses on particularly challenging cases, and \ourmethod{} significantly outperforms other baselines.
Please visit our project page for video results: \textcolor{blue}{\url{https://hifi-diffusion.github.io}}
\end{abstract}

\section{Introduction}

\begin{figure*}[!ht]
\centering
\footnotesize
\setlength\tabcolsep{1pt}
\newcommand{\teaserimgoverlay}[1]{\includegraphics[width=0.296\textwidth]{figures/teaser/#1_overlay.png}}
\newcommand{\teaserimg}[3]{\raisebox{-.83\height}{\includegraphics[width=0.167\textwidth]{figures/teaser/#1_#2#3.png}}}
\newcommand{\teaserrow}[1]{
\multirow{2}{*}{\teaserimgoverlay{#1}} & \teaserimg{#1}{m2m}{1} & \teaserimg{#1}{ema}{1} & \teaserimg{#1}{ours}{1} & \teaserimg{#1}{gt}{1}\\[1em]
& \teaserimg{#1}{m2m}{2} & \teaserimg{#1}{ema}{2} & \teaserimg{#1}{ours}{2} & \teaserimg{#1}{gt}{2}}
\begin{tabular}{c @{\hskip 0.5em} cccc}
    \teaserrow{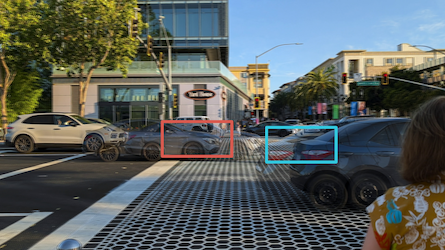} \\[3.9em]
    \teaserrow{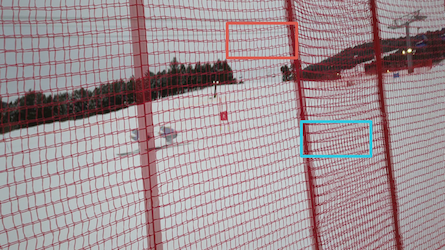} \\[3.9em]
    \teaserrow{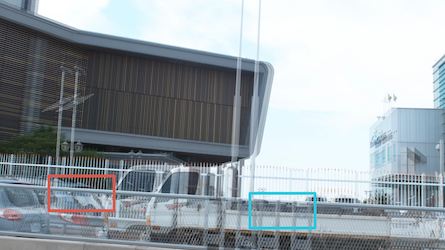} \\[3.9em]
    Input overlay & M2M & EMA-VFI & \textbf{\ourmethod{} (Ours)} & Ground Truth \\
\end{tabular}
\caption{
    \textbf{Qualitative comparison on challenging cases} on our proposed \ourdataset{} dataset (rows 1 and 2) and X-TEST (row 3). For challenging cases, such as large motion or repetitive textures, the proposed \ourmethod{} substantially outperforms other baselines.
    }
\label{fig:teaser}
\figureaftercaption
\end{figure*}

In a short amount of time, smartphone cameras have become both ubiquitous and significantly higher quality, capturing spatially higher resolution images and videos.
However, the temporal resolution---\ie video frame rate---of captured videos has lagged behind the spatial resolution, due to a combination of computational and memory costs and limited exposure time. 
The conflict between increased user interest in creative video content and technical limitations for capturing high frame-rate video has increased interest in techniques for high-resolution frame interpolation, which enables the synthesis of new frames between existing ones to enhance a video's frame rate.
Despite the progress, the latest techniques struggle in the high resolution setting, where challenging cases such as repetitive textures, detailed or thin objects become more common place.

Existing methods often design models using strong domain knowledge, \eg, correspondence matching~\cite{Ilg:2016:Flownet2,sun2018pwc,teed2020raft} and synthesis based on warping~\cite{Jiang:2018:SSM,Niklaus:2020:SSF,Park:2021:ABM,Xue:2019:VEW}. Domain knowledge enables small models to perform well when trained on a small amount of data but may restrict their capabilities. For example, when motion cues are incorporated into the model, the final quality are bounded by the accuracy of the motion. This is particularly evident on high resolution inputs with large motion, repetitive texture, and thin structures, where motion estimation often struggles (see \cref{fig:teaser}).

To address these challenges, we advocate a domain-agnostic diffusion approach, relying on model capacity and training data at scale for performance gains and generalization.
Some recent work have explored diffusion for frame interpolation but towards generative aspect, \eg better perceptual quality~\cite{Danier:2024:LDM} or complex and non-linear motion~\cite{Jain:2024:VID} between two frames very further apart in time.
Their performance, however, falls behind in the classical setting which predicts an intermediate frame \camready{and evaluates its fidelity to the ground truth using standard metrics, \eg, PSNR or SSIM.}

We instead introduce a \emph{\ourcascade{}} cascaded pixel diffusion approach for \textbf{Hi}gh resolution \textbf{F}rame \textbf{I}nterpolation, dubbed  \ourmethod{}. \ourmethod{} generalizes across diverse resolutions up to 8K images, a wide range of scene motions, and a broad spectrum of challenging scenes.
The diffusion framework allows us to scale both the model capacity and data size.
We also show that our model can effectively utilize large-scale video datasets. 
While cascades offer significant benefits for processing diverse input resolutions with different levels of motion, standard cascades, which denoise the entire high-resolution image, often struggle with memory issues at very high resolutions such as 8K.
To save memory during inference, we propose a new \emph{\ourcascade{}} cascade for frame interpolation, which always denoises the same resolution but is applied to patches of high resolution frames.
This also allows us to use one model for both base and super-resolution tasks, saving time for training separate models for both tasks and disk space at inference time.

The proposed \ourmethod{} method achieves state-of-the-art accuracy on challenging high-resolution public benchmark datasets, Xiph~\cite{Niklaus:2020:SSF}, X-TEST~\cite{Sim:2021:XVF} and SEPE~\cite{al2023sepe}, and demonstrates strong performance on challenging corner cases, \eg, repetitive textures and large motion. We also introduce a new evaluation dataset, \ourdatasetlong{} (\ourdataset{}),
which specifically highlights these challenging cases and demonstrate that \ourmethod{} significantly outperforms existing baselines.

\section{Related Work}

\paragraph{Domain-specific architecture for interpolation.}
Motion-based approaches synthesize intermediate frames using estimated bi-directional optical flow between two nearby frames.
These methods employ forward splatting \cite{Hu:2022:M2M,Jin:2023:UPR,Niklaus:2018:CAS,Niklaus:2020:SSF} or backward warping \cite{Huang:2022:RIF,Jiang:2018:SSM,Park:2021:ABM,Park:2020:BMB,Sim:2021:XVF}, followed by a refinement module that improves visual quality. Performance is often bounded by motion estimation accuracy, as inaccuracies in the motion cause artifacts during the splatting or warping process.
As a result, they struggle on inputs for which optical flow estimation is problematic, \eg,~large motion, occlusion, and thin objects. 

Phase-based approaches \cite{Meyer:2015:PBF,Meyer:2018:PNV} propose to estimate an intermediate frame in a phase-based representation instead of the conventional pixel domain.
Kernel-based approaches \cite{Cheng:2020:VFI,Lee:2020:ACA,Niklaus:2021:RAC,Niklaus:2017:VAC,Niklaus:2017:VSC} present simple single-stage formulations that estimate per-pixel $n\times n$ kernels and synthesize the intermediate frame using convolution on input patches.
Both approaches avoid reliance on motion estimator, but they do not usually perform well on high resolution input with large motion, even with deformable convolution~\cite{Cheng:2020:VFI}.

\paragraph{Generic architecture for interpolation.}
Some methods explore using a generic architecture without domain knowledge, such as attention~\cite{Choi:2020:CAI},  transformer~\cite{Shi:2022:VFI}, 3D convolution under multi-frame input setup \cite{Kalluri:2023:FLA,Shi:2022:VFI}. 
However, both attention and 3D convolution are computationally expensive and thus prohibitive at 4K or 8K resolution.

\paragraph{Diffusion models for computer vision.}
Recently diffusion models have demonstrated their strength on generative computer vision applications such as image~\cite{Ho:2022:CDM,Peebles:2023:SDM} and video generation~\cite{Blattmann:2023:SVD,Ho:2022:VDM}, image editing~\cite{Brooks:2023:IPP,Yang:2023:ZCL}, 3D generation~\cite{Qian:2024:MOT}, \etc.
Beyond generation, diffusion has also shown to be effective for dense computer vision tasks and has been become the state-of-the-art technique for classical problems such as depth prediction~\cite{Ke:2024:RDI,Saxena:2023:TSE}, optical flow prediction~\cite{Saxena:2023:TSE}, correspondence matching~\cite{Nam:2024:DMD}, semantic segmentation \cite{Baranchuk:2022:LES,Xu:2023:OPS}, \etc.

\paragraph{Diffusion models for interpolation.}
Two recent works explore diffusion for video frame interpolation from a generative perspective.
LDMVFI~\cite{Danier:2024:LDM} proposes using a conditional latent diffusion model and optimizes it for perceptual frame interpolation quality, but the PSNR or SSIM metric of the predicted frames tends to be lower than that by state of the art.
VIDIM~\cite{Jain:2024:VID} uses a cascaded pixel diffusion model but focuses on a task closer to the conditional video generation. Given two temporally-far-apart frames, the method generates a base video of 7 frames at $64 \times 64$ resolution and then upsamples them to $256 \times 256$.
It is unclear whether a diffusion-based approach can achieve competitive results on the classical frame interpolation problem, where the input frames come from a video with high FPS and can be up to 8K resolution.

\paragraph{Cascaded diffusion models.}
Beginning with CDM~\cite{Ho:2022:CDM}, cascades have become standard for scaling up the output resolution of pixel diffusion models.
Diffusion cascades consist of a ``base'' model for an initial low-resolution solution and a number of separate ``super-resolution'' models to produce a higher-resolution output conditioned on the low resolution output.
While effective for high-resolution output, memory cost increases proportionally with resolution since each super-resolution model performs diffusion at its output resolution.
Even with specialized super-resolution architectures~\cite{Ho:2022:CDM,Saharia:2022:PTD}, the memory problem still persists as the target resolution increases significantly, \eg~from 1K to 8K.

\paragraph{High resolution diffusion.}
Recent works in high-resolution image generation have introduced training-free approaches through merging the score functions of nearby patches~\cite{bar2023multidiffusion,liu2023syncdreamer} or expanding the network~\cite{shi2024resmaster, kim2024diffusehigh}.
Other methods explicitly train models to denoise partitioned patches (or tiles) and then merge them into high-resolution output.
\citet{zheng2024any} generates any-size high-resolution output by merging denoised non-overlapping tiles during sampling process. 
\citet{ding2023patched} uses score value and feature maps to encourage consistency between denoised patches.
\citet{skorokhodov2024hierarchical} uses a hierarchical patch structure for efficient video generation, but requires specialized modules for global consistency.

Unlike these efforts, we focus on frame interpolation, an estimation task, and improve inference memory efficiency at extremely high resolutions (4K or 8K). Generation tasks require inter-patch communication for coherence generation at high resolution.
Estimation task, however, benefits from strong conditioning signals (input frames) that localize the problem at the patch level.
This allows us to explore distinct architectural choices for frame interpolation.

For estimation tasks, the most similar to ours is DDVM~\cite{Saxena:2023:TSE} which uses tiling for high-resolution inference.
After the base model runs at a coarse solution, the output is upsampled by partially denoising tiles taken from the coarse solution and input frames.
In the context of frame interpolation, we show that this tiling performs worse than our proposed \ourcascade{} cascade.

\section{Diffusion for High-Resolution Frame Interpolation}

We propose a pixel diffusion approach for frame interpolation.
To enable high-resolution inference with low memory usage, we introduce a novel cascading strategy, \ourcascade{}, which performs diffusion on patches of high-resolution inputs.
This cascade allows us to use the same model for both base estimation and upsampling.

\subsection{Architecture}\label{sec:method_base}

\begin{figure}[t]
\centering
\includegraphics[width=\linewidth]{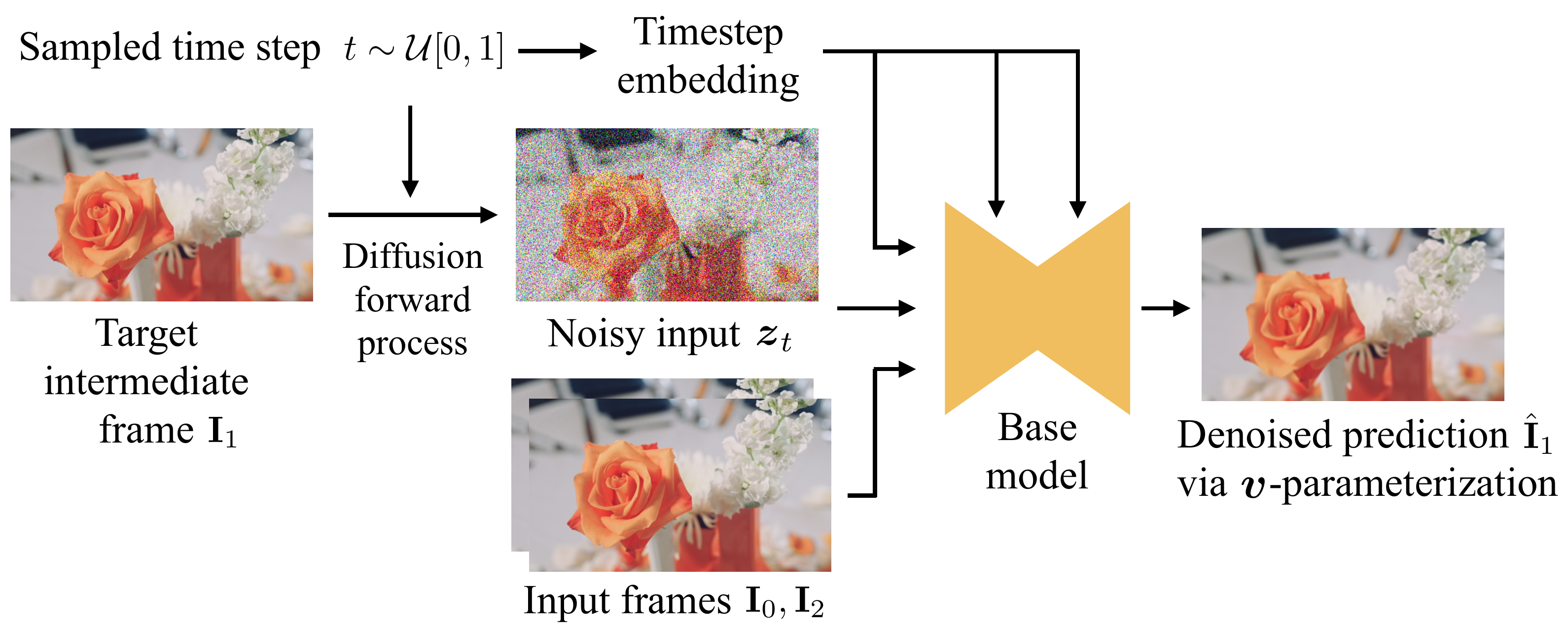} 
\caption{Our \textbf{base model} is conditioned on two input frames, $\mathbf{I}_0$ and $\mathbf{I}_2$, and predicts the intermediate frame $\mathbf{I}_1$. The model uses $\bm{v}$-parameterization \cite{Salimans:2022:PDF,Saxena:2023:ZSM} for both model output and loss.
}
\label{fig:base_model}
\figureaftercaption
\end{figure}

Our method adopts a conditional image diffusion framework.
Given a concatenation of temporally nearby frames $\mathbf{I}_0$ and $\mathbf{I}_2$ as a conditioning signal, our model aims to estimate an intermediate frame $\hat{\mathbf{I}}_1$ as a reverse diffusion process in the pixel space, as illustrated in \cref{fig:base_model}.
We take a generic efficient U-Net architecture from DDVM \cite{Saxena:2023:TSE} with $\bm{v}$-parameterization \cite{Salimans:2022:PDF,Saxena:2023:ZSM} for both model output and loss.
The U-Net includes self-attention layers at two bottom levels.
Given a noisy image $\bm{z}_t = \alpha_t \bm{x} + \sigma_t \bm{\epsilon}$ as an input, the network predicts $\hat{\bm{v}}$, where $\bm{x}$ is the target image (\ie, $\mathbf{I}_1$), sampled random noise $\bm{\epsilon} \sim \mathcal{N}(\bm{0}, \bm{I})$, sampled time step $t \sim \mathcal{U}[0, 1]$, and $\alpha^2_t + \sigma^2_t = 1$.
We directly apply L1 loss on $\bm{v}$ parameter space, \ie, $||\hat{\bm{v}} - \bm{v}||_1$, where $\bm{v} = \alpha_t \bm{\epsilon} - \sigma_t \bm{x}$.
The predicted image is recovered by $\hat{\bm{x}} = \alpha_t \bm{z}_t - \sigma_t \hat{\bm{v}}$, where $\hat{\bm{x}} = \hat{\mathbf{I}}_1$.

\subsection{\ourCascade{} cascade model}\label{sec:method_cascade}

\begin{figure*}[t]
\centering
\includegraphics[trim={0 8pt 0 8pt},clip,width=\linewidth]{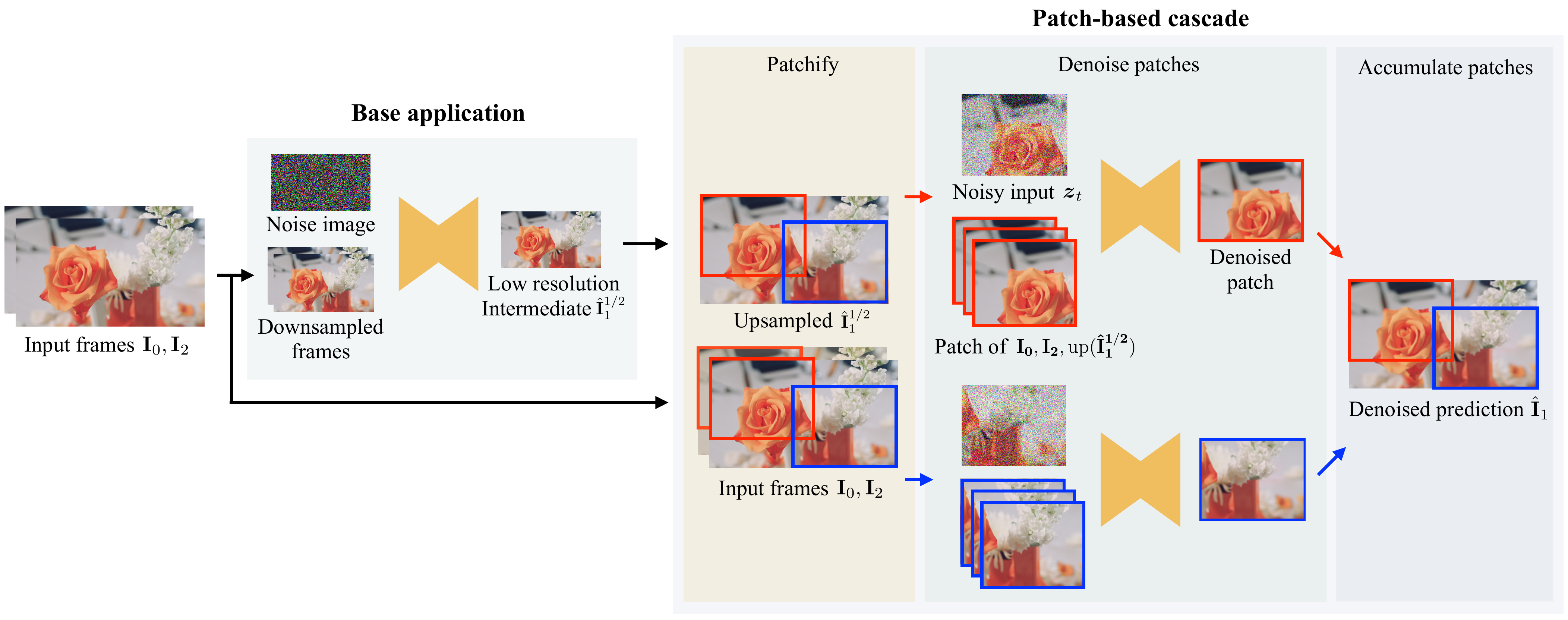}
\caption{
\textbf{\ourCascade{} cascade model}. 
Given a low-resolution intermediate from the previous level, \ourcascade{} cascade creates patches from bi-linearly upsampled low-resolution intermediate and two input frames and uses these patches as conditioning for a diffusion process. It then combines denoised patches to form the whole image. At inference time, only a single weight-shared model is recursively used across different image scales as in~\cref{fig:overall}. Two-stage cascade is shown for simplicity.
}
\label{fig:arch}
\figureaftercaption
\end{figure*}

\begin{figure}[t]
\centering
\includegraphics[width=0.95\linewidth]{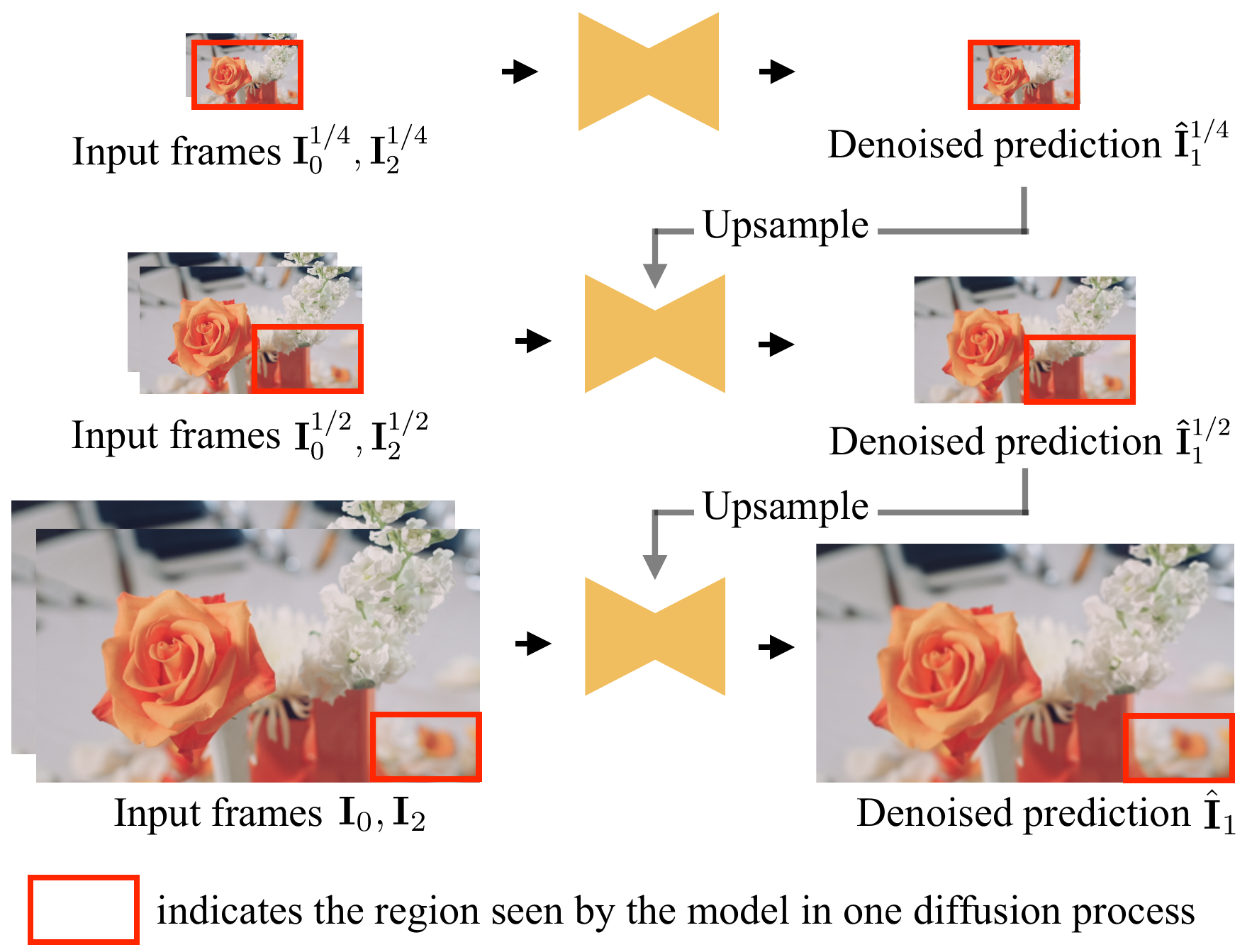}
\caption{
\textbf{Upsampling strategy}. Like a standard cascade, we process the image from coarse to fine, but we always denoise at the same resolution, as indicated by the red box. Details on each step of the cascade are in~\cref{fig:arch}.
}
\label{fig:overall}
\figureaftercaption
\end{figure}

To handle extremely high resolutions (up to 8K), we propose a \ourcascade{} cascade approach that performs diffusion on patches of the input frames, keeping peak memory usage near constant at inference time. This allows us to use the same architecture for every upsample level and reuse the same model for both base and super-resolution settings, saving training time and disk space.
On such high resolutions, standard cascades, which denoise the entire high resolution image directly, would require either a considerable amount of memory at inference time or a careful architecture search to reduce the memory cost.

\paragraph{Approach.} 
\cref{fig:overall} shows our overall \camready{inference} strategy: we adopt the well-known coarse-to-fine idea for cascades and build an $N$-level image pyramid. 
Starting from the lowest scale $s_{N-1}$ (where $s_n \!\equiv\! \nicefrac{1}{2^{n}}$), we downsample the input conditioning images by a factor of $s_{N-1}$ (\ie, $\mathbf{I}^{s_{N-1}}_{0}$ and $\mathbf{I}^{s_{N-1}}_{2}$) and  predict an intermediate image $\hat{\mathbf{I}}^{s_{N-1}}_1$ at the same scale $s_{N-1}$.
We then apply $2 \times$ bilinear upsampling to this intermediate image $\text{up}(\mathbf{\hat{I}}^{s_{N-1}}_1)$ and use it as a conditioning signal for a denoising process at the scale $s_{N-2}$.

At each pyramid level, we upscale prediction via \ourcascade{} cascade, as shown in~\cref{fig:arch}. For refinement at scale $s_{N-2}$, we take the prediction from the prior scale and then upsample it to match $s_{N-2}$. At each level, we perform three stages: \emph{(i)} patchify, where we crop overlapping patches from the upsampled intermediate prediction and the input at that scale, \emph{(ii)} denoise patches, where we run diffusion to obtain the prediction for each patch,
and \emph{(iii)} accumulate patches, where we use MultiDiffusion~\cite{bar2023multidiffusion} to merge results from different patches for the prediction $\mathbf{\hat{I}}^{s_{N-2}}_{2}$.
\camready{Here, we merge denoised patches at every denoising step.}
We then upsample $\mathbf{\hat{I}}^{s_{N-2}}_{2}$ to level $s_{N-3}$ and repeat this process until $n\!=\!0$, the original input scale.

\paragraph{Training setup.} 
For the \ourcascade{} cascade model, we want to train a diffusion model that is conditioned on a pair of input images and a half resolution representation of the target we aim to predict. We first predict the intermediate frame at a half resolution $\mathbf{\hat{I}}_1^{\nicefrac{1}{2}}$ by feeding downsampled inputs to a pre-trained base model (\cref{fig:base_model}) computed offline. 
This intermediate frame is then upsampled to the original scale and used as a conditioning image, along with the original inputs, for training the \ourcascade{} cascade model using standard diffusion.
This inference step is performed offline to improve training efficiency.

\begin{figure*}[!t]
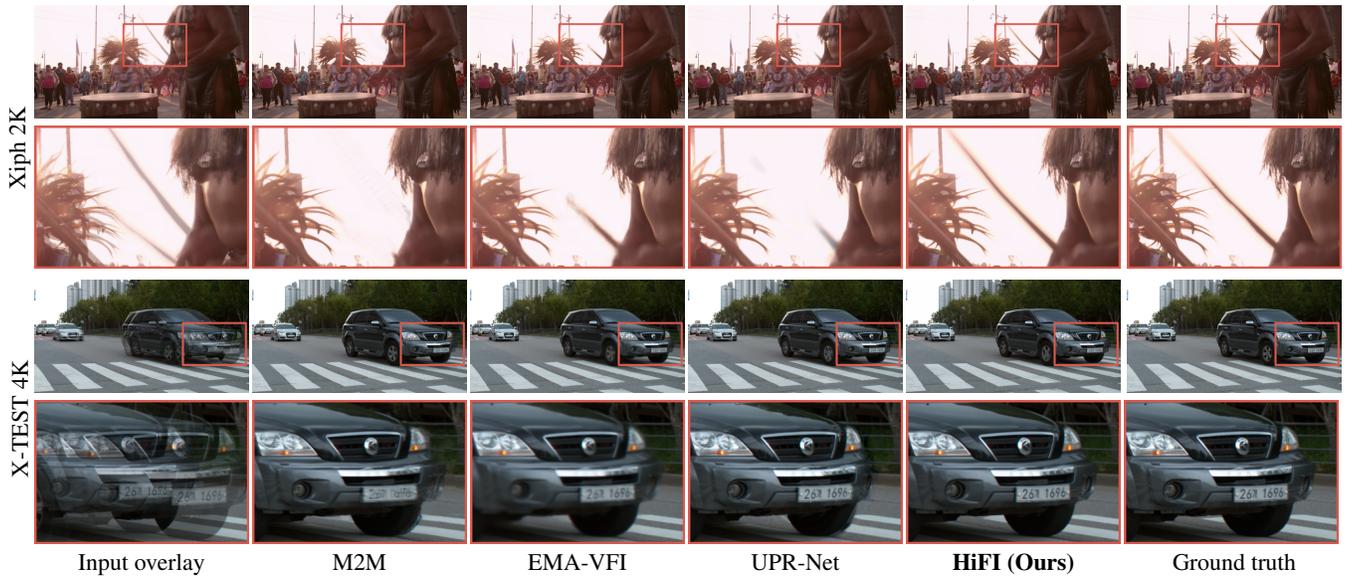

\centering
\footnotesize
\setlength\tabcolsep{0.75pt}
\newcommand{\qualimg}[4]{\includegraphics[width=0.16\textwidth]{figures/qual/#1/#2_#3_#4.png}}
\newcommand{\qualrow}[3]{
\qualimg{#1}{#2}{#3}{overlay} & 
\qualimg{#1}{#2}{#3}{m2m} & 
\qualimg{#1}{#2}{#3}{ema} &
\qualimg{#1}{#2}{#3}{upr} &
\qualimg{#1}{#2}{#3}{ours} &
\qualimg{#1}{#2}{#3}{gt}
} 

\begin{tabular}{ccccccc}
    & \qualrow{xiph2k}{ritualdance_074_0055}{full}\\
    \raisebox{30pt}[0pt][0pt]{\rotatebox{90}{\footnotesize Xiph 2K}} & \qualrow{xiph2k}{ritualdance_074_0055}{crop} \\[0.15em]
    & \qualrow{x4k}{car}{full}\\
    \raisebox{25pt}[0pt][0pt]{\rotatebox{90}{\footnotesize X-TEST 4K}}
    & \qualrow{x4k}{car}{crop} \\
    & Input overlay & M2M & EMA-VFI & UPR-Net & \textbf{\ourmethod{} (Ours)} & Ground truth \\
\end{tabular}
    \caption{\textbf{Qualitative examples for public datasets.} Our method performs well even in cases of large motion and complex textures such as a thin object on the top and the plate number at the bottom. }
    \label{fig:qual_public}
    \figureaftercaption
\end{figure*}

\paragraph{Single model for all stages.} By conditioning on the low resolution estimate but using dropout 50\% during training time, we can use the same model for all cascade stage, including base and super-resolution; base generation is done by passing zeros as the low resolution condition. 
While smilar to CFG~\cite{ho2022classifier}, we do not combine unconditional and conditional estimations at inference.
Empirically we find that a single shared model for all stages performs slightly better than having a dedicated super-resolution model.
It also substantially reduces training time (training one model instead of multiple separate ones) and disk space at inference time (saving only one model).
Interestingly, we observe that a dedicated super-resolution model trained without dropout on the coarse estimation does not work since it takes the shortcut of upsampling the low resolution instead of attending to the high resolution inputs.

\section{Experiments}

\paragraph{Implementation details.}
Similar to previous diffusion-based methods \cite{Jain:2024:VID,Danier:2024:LDM}, we utilize a large-scale video dataset for training, to test the scalability of the diffusion model better.
The dataset contains up to 30~M videos with 40 frames, collected from the internet and other sources \camready{with licenses permitting research uses}.
We first train our base model on the dataset, and then we additionally include Vimeo-90K triplet~\cite{Xue:2019:VEW} and X-TRAIN~\cite{Sim:2021:XVF} to finetune the cascade model.
For fair comparison, we also prepare a model trained on Vimeo-90K and X-TRAIN only from scratch.
We use a mini-batch size of 256 and train the base model for 3~M iteration steps and the \ourcascade{} cascade model for 200~k iteration steps. We use the Adam optimizer \cite{Kingma:2014:ADA} with a constant learning rate $1e^{-4}$ with initial warmup.
For inference, we use 3-stage \ourcascade{} cascade setup with a patch size of $512\times768$, averaging 4 samples estimated via 4 sampling steps.

Our data augmentation includes random crop and horizontal, vertical, and temporal flip with a probability of 50\%.
We use a crop size of $352\times480$ for large-scale base model training and $224\times288$ for the cascade model training. 
We use a multi-resolution crop augmentation that crops an image patch with a random rectangular crop size between the original resolution and the final crop size and then resize it to the final crop size.
While commonly used, we find random 90$^{\circ}$ rotation augmentation and photometric augmentation to be less effective, so we opt not to use them.

More details are in the supplementary material.

\begin{table*}[!t]
\centering
\scriptsize
\setlength\tabcolsep{5.7pt}
\begin{tabularx}{\linewidth}{@{} X @{\hskip 2em} c @{\hskip 2em} *{5}{S[table-format=2.2] S[table-format=1.3]}}
\toprule
\multirow{3}{*}{Method} & \multirow{3}{*}{Training dataset} & \multicolumn{2}{c}{\multirow{2}{*}{Vimeo-90K}} & \multicolumn{4}{c}{Xiph} & \multicolumn{4}{c}{X-TEST} \\
& & & & \multicolumn{2}{c}{2K} & \multicolumn{2}{c}{4K} & \multicolumn{2}{c}{2K} & \multicolumn{2}{c}{4K} \\
\cmidrule(lr){3-4} \cmidrule(lr){5-6} \cmidrule(lr){7-8} \cmidrule(l){9-10} \cmidrule(l){11-12}
& & PSNR & SSIM & PSNR & SSIM & PSNR & SSIM & PSNR & SSIM & PSNR & SSIM \\
\midrule
M2M \cite{Hu:2022:M2M} & Vimeo & 35.47 & 0.978 & 36.44 & \secondbest 0.967 & 33.92 & 0.945 & 32.07 & 0.923 & 30.81 & 0.912 \\
FILM \cite{Reda:2022:FIL} & Vimeo & 36.06 & 0.970 & 36.66 & 0.951 & 33.78 & 0.906 & 31.61 & 0.916 & 26.98 & 0.839 \\
AMT \cite{Li:2023:AMT} & Vimeo  & 36.53 & \best 0.982 & 36.38 & 0.941 & 34.63 & 0.904 & {-} & {-} & {-} & {-} \\
UPR-Net \cite{Jin:2023:UPR} & Vimeo  & 36.42 & \best 0.982 & {-} & {-} & {-} & {-} & {-} & {-} & 30.68 & 0.909 \\
FITUG \cite{Plack:2023:FIT} & Vimeo  & 36.34 & \secondbest 0.981 & {-} & {-} & {-} & {-} & {-} & {-}  & {-} & {-} \\ 
TCL \cite{Zhou:2023:EMA} & Vimeo  & \best 36.85 & \best 0.982 & {-} & {-} & {-} & {-} & {-} & {-} & {-} & {-} \\ 
IQ-VFI \cite{Hu_2024_CVPR} & Vimeo & 36.60 & \best 0.982 & 36.68 & 0.942 & \secondbest 34.72 & 0.905 & {-} & {-} & {-} & {-} \\
EMA-VFI \cite{Zhang:2023:EMA} & Vimeo \camready{(+septuplet for X-TEST)} & \secondbest 36.64 & \best0.982 & \secondbest 36.90 & 0.945 &  34.67 & 0.907 & 32.85 & \secondbest 0.930 & 31.46 & 0.916 \\
XVFI \cite{Sim:2021:XVF} & Vimeo / X-TRAIN & 35.07 & 0.976 & {-} & {-} & {-} & {-} & 30.85 & 0.913 & 30.12 & 0.870 \\
BiFormer~\cite{Park:2023:BFL} & Vimeo + X-TRAIN & {-} & {-} & {-} & {-} & 34.48 & 0.927 & {-} & {-} & 31.32 & \secondbest 0.921\\
\midrule
\multirow{2}{*}{\textbf{\ourmethod{} (Ours)}}
 & Vimeo + X-TRAIN & 35.70 & 0.979 & 36.64 &  \secondbest 0.967 & 34.45 & \secondbest 0.948 & \secondbest 33.03 & 0.927 & \secondbest 32.03 & 0.918 \\
 & Vimeo + X-TRAIN + Raw videos & 36.12 & 0.980 & \best 37.36 & \best 0.969 &  \best 35.40 & \best 0.953 & \best 33.94 & \best 0.941 & \best 32.92 & \best 0.931 \\
 
\bottomrule
\end{tabularx}
\caption{\textbf{Results on public benchmark datasets}: \ourmethod{} performs favorably on the highly-saturated Vimeo-90K~\cite{Xue:2019:VEW} and is substantially more accurate than existing two-frame methods on high-resolution Xiph~\cite{Niklaus:2020:SSF} and X-TEST~\cite{Sim:2021:XVF} datasets. 
\colorbox{bestcol}{\textbf{Best}} and \colorbox{secondbestcol}{second-best} are highlighted in color.}
\label{tab:state_of_the_art_table}
\tablebelow
\end{table*}

\subsection{Public benchmark evaluation}

We first evaluate \ourmethod{} on three
popular benchmark datasets, Vimeo-90K triplet~\cite{Xue:2019:VEW}, Xiph~\cite{Niklaus:2020:SSF}, and X-TEST~\cite{Sim:2021:XVF} in \cref{tab:state_of_the_art_table}, as well as an 8K dataset, SEPE~\cite{al2023sepe}.

\paragraph{Vimeo-90K.}
The low-resolution  ($256\!\times\!448$) Vimeo-90K is one of the most heavily studied benchmark, where numbers are highly saturated among different methods.
\ourmethod{} achieves competitive accuracy with a generic training procedure. 
Please view the supplementary for further discussion.

\paragraph{Xiph and X-TEST.}
Both Xiph and X-TEST have high resolution (2K and 4K).  The motion of X-TEST can be over 400 pixels at the 4K resolution, particularly challenging for existing methods. 
For X-TEST, we follow the evaluation protocol discussed in~\cite{Sim:2021:XVF} that interpolates 7 intermediate frames.
When trained on a combination of Vimeo and X-TRAIN, \ourmethod{} performs favorably against state of the art on Xiph and X-TEST datasets, both in 2K and 4K resolutions.
Pre-training on a large video dataset significantly boosts the performance of \ourmethod{} on Xiph and X-TEST, setting a new state of the art.
Visually, \ourmethod{} can better interpolate fine details with large motion at high resolution, as shown in \cref{fig:qual_public}. 
We will analyze key components that contribute to the performance in the ablation study below. 

\paragraph{SEPE.}
We also test \ourmethod{} on SEPE that includes 8K resolution videos.
Most methods we tested ran out of memory except \camready{M2M (PSNR 28.34 (dB) and SSIM 0.883) and SGM-VFI~\cite{Liu_2024_CVPR} (PSNR 28.43 (dB) and SSIM 0.880),} compared with PSNR 29.78 (dB) and SSIM 0.900 by \ourmethod{}. Please view the supplementary for visual comparison.

\begin{figure}[!t]
\centering
\footnotesize
\setlength\tabcolsep{0.4pt}
\newcommand{\datasetimgwidth}{0.325\linewidth}
\begin{tabular}{c @{\hskip 0.3em} c @{\hskip 0.3em} c}
	\includegraphics[width=\datasetimgwidth]{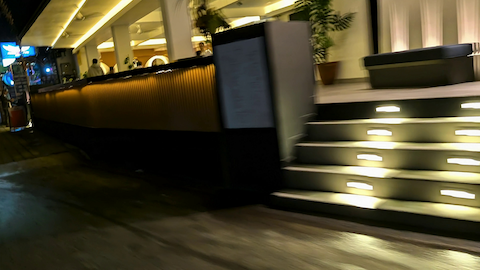} & 
	\includegraphics[width=\datasetimgwidth]{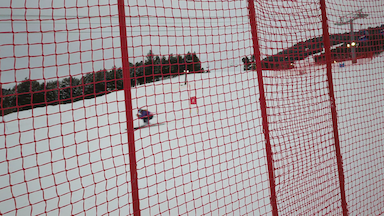} & 
	\includegraphics[width=\datasetimgwidth]{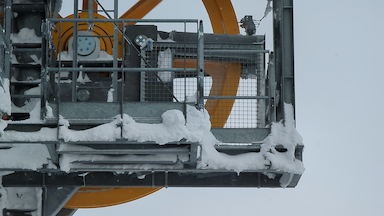} \\
	\includegraphics[width=\datasetimgwidth]{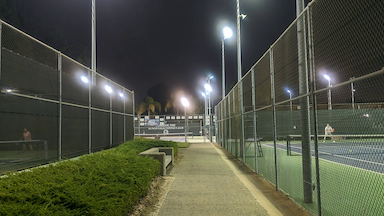} & 
	\includegraphics[width=\datasetimgwidth]{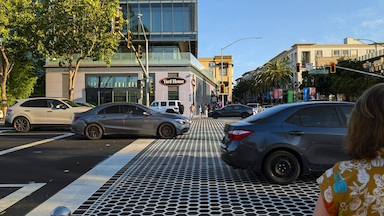} & 
	\includegraphics[width=\datasetimgwidth]{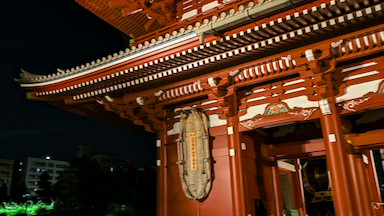} \\
\end{tabular}
\caption{
\textbf{A few examples from our \ourdataset~dataset} that includes challenging scenes, such as repetitive texture and large motion where typical methods fail.}
\label{fig:dataset_example}
\end{figure}

\subsection{\ourdatasetlong{} dataset}

Public benchmark datasets, while diverse, do not fully capture the failure modes of current methods, especially large motion or repetitive texture cases common in real-world videos.
To better evaluate existing methods and further innovation, we introduce a \ourdatasetlong{} (\textbf{\ourdataset{}}) dataset that includes such 19 challenging examples at 4K resolution in both portrait and landscape modes, as shown in \cref{fig:dataset_example}. 
As in \cref{tab:benchmark_custom_data} and \cref{fig:qual_lamor}, \ourmethod{} substantially outperform all state of the arts on challenging cases of repetitive textures and large motion.

\begin{table}[!t]
\centering
\scriptsize
\setlength\tabcolsep{4.5pt}
\begin{tabularx}{0.8\linewidth}{@{} X @{\hskip 2em} *{2}{S[table-format=2.3]}}
\toprule
Method & PSNR & SSIM\\
\midrule
{LDMVFI~\cite{Danier:2024:LDM}} & 21.952 & 0.828 \\
{EMA-VFI~\cite{Zhang:2023:EMA}} & 22.327 & 0.845 \\
{M2M~\cite{Hu:2022:M2M}} & 24.995 & 0.884 \\
{SGM-VFI~\cite{Liu_2024_CVPR}} & 25.122 & \secondbest 0.894 \\
{UPR-Net~\cite{Jin:2023:UPR}} & 25.856 & 0.892 \\
{BiFormer~\cite{Park:2023:BFL}} & \secondbest 26.330 & 0.893 \\
{\textbf{\ourmethod{} (Ours)}} & \best 28.141 & \best 0.912 \\
 
\bottomrule
\end{tabularx}
\caption{Results on \textbf{\ourdataset{}.} \ourmethod{} is significantly more accurate than state-of-the-art methods.}
\label{tab:benchmark_custom_data}
\tablebelow
\end{table}

\begin{figure*}[!ht]
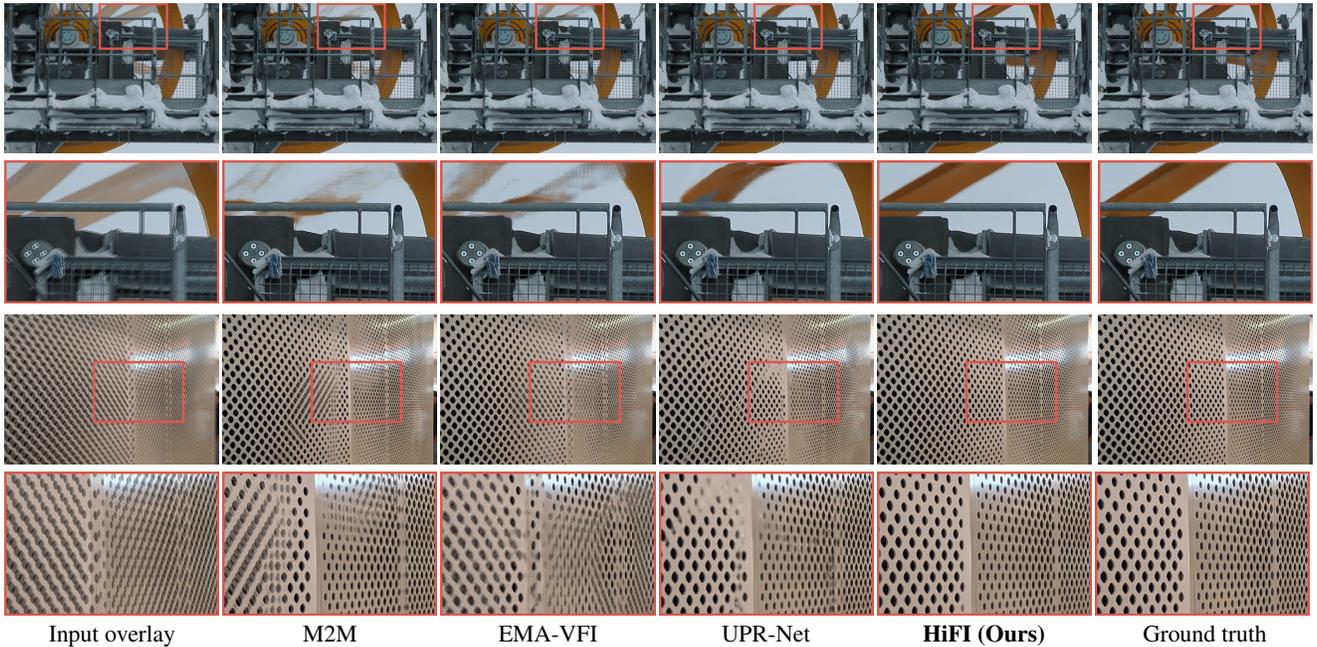

\centering
\footnotesize
\setlength\tabcolsep{0.75pt}
\newcommand{\qualimg}[4]{\includegraphics[width=0.16\textwidth]{figures/qual/#1/#2_#3_#4.png}}
\newcommand{\lamorqualrow}[3]{
\qualimg{#1}{#2}{#3}{overlay} & 
\qualimg{#1}{#2}{#3}{m2m} & 
\qualimg{#1}{#2}{#3}{ema} &
\qualimg{#1}{#2}{#3}{upr} &
\qualimg{#1}{#2}{#3}{ours} &
\qualimg{#1}{#2}{#3}{gt}
} 

\begin{tabular}{cccccc}
    \lamorqualrow{lamor}{rolling}{full}\\
    \lamorqualrow{lamor}{rolling}{crop} \\[0.15em]
    \lamorqualrow{lamor}{holes}{full}\\
    \lamorqualrow{lamor}{holes}{crop} \\
    Input overlay & M2M & EMA-VFI & UPR-Net & \textbf{\ourmethod{} (Ours)} & Ground truth \\
\end{tabular}
    \caption{\textbf{Qualitative comparison on \ourdataset{}.} The proposed \ourmethod{}
    is particularly effective at very challenging cases including repetitive textures and large motion. 
    }
    \label{fig:qual_lamor}
    \vspace{-1em}
\end{figure*}

\subsection{Ablation study}
\label{sec:ablation}

\paragraph{Dedicated upsample model vs single model.} In  \cref{tab:cascade_comparison}, we compare the accuracy of the base model, two distinct models for base and upsample, and our final setting of using the same model for base and upsample. 
Both cascade strategies are effective for handling large motion, substantially improving accuracy on X-TEST.
Using the same model for both base and upsample performs on-par or even better than having a dedicated upsample model, especially on challenging X-TEST.
This validates the strength of re-using the same model for both over the more expensive dedicate model setup.
Increasing the number of upsample stages improves the accuracy but saturates over three.

\begin{table}[!t]
\centering
\scriptsize
\setlength\tabcolsep{1.4pt}
\begin{tabularx}{\linewidth}{@{} X c @{\hskip 1em} *{3}{S[table-format=2.2] S[table-format=1.3]} @{}}
\toprule
\multirow{2}{*}{Method} & \multirow{2}{*}{Model size} &\multicolumn{2}{c}{Vimeo-90K} & \multicolumn{2}{c}{X-TEST 2K} & \multicolumn{2}{c}{X-TEST 4K} \\
\cmidrule(lr){3-4} \cmidrule(lr){5-6} \cmidrule(lr){7-8}
& & PSNR & SSIM & PSNR & SSIM & PSNR & SSIM\\
\midrule
Base & 647 M & \secondbest 35.44 & \secondbest 0.978 & 30.32 & 0.879 & 28.57 & 0.876 \\ 
\midrule
Standard two-stage cascade & 1294 M & \best 36.12 & \best 0.980 & 33.86 & 0.939 & 32.48 & 0.926 \\
\ourCascade{} self-cascade $\times$ 2 & 647 M & \best 36.12 & \best 0.980 & \secondbest 33.93 & \best 0.941 & \secondbest 32.77 & \secondbest 0.930 \\
\ourCascade{} self-cascade $\times$ 3 & 647 M & \best 36.12 & \best 0.980 & \best 33.94 & \best 0.941 & \best 32.92 & \best 0.931 \\
\bottomrule
\end{tabularx}
\caption{\textbf{Base \vs cascade models.} Our \ourcascade{} self-cascade formulation substantially outperforms the base with the same number of parameters.
Through model sharing, our self-cascade generalizes better on the X-TEST dataset than the standard cascade but with half of the parameters.}
\label{tab:cascade_comparison}
\tablebelow
\end{table}

\paragraph{Comparison with coarse-to-fine refinement.}
Coarse-to-fine tiling refinement from DDVM \cite{Saxena:2023:TSE} first predicts the target at low resolution, bilinearly upsamples it to the target resolution, and refines it from an intermediate sampling step in a patch-wise manner. Our \ourcascade{}  cascade performs consistently better than the coarse-to-fine tiling refinement on the X-TEST benchmark; 32.92 (dB) vs 32.54 (dB) on 4K, and 33.94 (dB) \vs 33.03 (dB) on 2K.

\paragraph{Architecture.}
In \cref{tab:model_analysis}, we analyze where the major gain originates from by ablating attention layers or diffusion process in the base model, given the same training assets (\eg, datasets, computations, \etc). 
Using attention layers brings about moderate performance gains on both the small (Vimeo) and large (X-TEST) motion datasets. 
\camready{We find the attention layers help with handling large motion and repetitive textures, enabling the accurate interpolation of frames by capturing the global context of these textures.}
Removing the diffusion process also leads to significant performance degradation. 
We also test one widely-used traditional method FILM~\cite{Reda:2022:FIL}, which relies on a ``scale-agnostic'' motion estimator to handle large motion. FILM trained on the same large dataset is substantially worse than \ourmethod{}, suggesting that traditional, hand-designed methods do not scale up well \wrt data.

\begin{table}[!t]
\centering
\scriptsize
\setlength\tabcolsep{3pt}
\begin{tabularx}{\linewidth}{@{} X @{\hskip 2em} *{3}{S[table-format=2.2] S[table-format=1.3]} @{}}
\toprule
\multirow{2}{*}{Method} & \multicolumn{2}{c}{Vimeo-90K} & \multicolumn{2}{c}{X-TEST 2K} & \multicolumn{2}{c}{X-TEST 4K} \\
\cmidrule(lr){2-3} \cmidrule(lr){4-5} \cmidrule(lr){6-7}
& PSNR & SSIM & PSNR & SSIM & PSNR & SSIM\\
\midrule
\textbf{Ours}, base model & \best 35.44 & \best 0.978 & \best 30.32 & \best 0.879 & \best 28.57 & \best 0.876 \\
 \; w/o attention layers & \secondbest 35.13 & \secondbest 0.977 & \secondbest 29.73 & \secondbest 0.861 & \secondbest 27.75 & 0.854 \\
 \; w/o diffusion & 33.78 & 0.965 & 28.05 & 0.852 & 27.56 & \secondbest 0.861 \\
\midrule   
FILM \cite{Reda:2022:FIL} & 34.02 & 0.970 & 28.15 & 0.854 & 27.24 & 0.856 \\
\bottomrule
\end{tabularx}
\caption{\textbf{Architecture analysis.} Both attention layers and diffusion process contribute to substantial accuracy gain. Comparing to a domain-specific architecture, FILM, our approach scales up better when training on the same large-scale video dataset.}
\label{tab:model_analysis}
\tablebelow
\end{table}

\paragraph{Number of sampling steps.}
The optimal number of sampling steps also differ between the base and the \ourcascade{} cascade model.
In general, we find that the model needs more sampling steps for large motion (\eg, X-TEST) than for small motion (\eg, Vimeo-90K \cite{Xue:2019:VEW}).
However, the \ourcascade{} cascade model is able to achieve better numbers across different datasets with fewer sampling steps.

\begin{table}[!t]
\centering
\scriptsize
\begin{subtable}{.48\linewidth}
\centering
\begin{tabularx}{0.93\linewidth}{@{} X @{\hskip 2em} S[table-format=2.2,round-mode=places,round-precision=2] S[table-format=2.2,round-mode=places,round-precision=2]}
\toprule 
{Steps} & {Vimeo-90K} & {X-TEST 4K} \\
\midrule
1 & 34.87 & 27.95 \\
2 & \secondbest 35.37 & 27.92 \\
4 & \best 35.44 & 28.57 \\
8 & 35.21 & 29.67 \\
16 & 34.58 & \secondbest 30.34 \\
32 & 33.53 & \best 30.40 \\
64 & 32.68 & 30.02 \\
\bottomrule
\end{tabularx}
\caption{\scriptsize Base model}
\end{subtable}
\begin{subtable}{.48\linewidth}
\centering
\begin{tabularx}{0.93\linewidth}{@{} X @{\hskip 2em}
S[table-format=2.2,round-mode=places,round-precision=2] S[table-format=2.2,round-mode=places,round-precision=2]}
\toprule
{Steps} & {Vimeo-90K} & {X-TEST 4K} \\
\midrule
1 & \secondbest 36.13 & 32.32 \\
2 & \best 36.15 & 32.83 \\
4 & 36.12 & \best 32.92 \\
8 & 36.06 & \best 32.92 \\
16 & 35.98 & 32.84 \\
32 & 35.92 & 32.68 \\
64 & 35.86 & 32.64 \\
\bottomrule
\end{tabularx}
\caption{\scriptsize \ourCascade{} cascade}
\end{subtable}

\caption{\textbf{Effect of the sampling steps} on PSNR for the base model and \ourcascade{} cascade model. More steps tends to be better for higher resolution and large motion datasets. } 
\label{tab:denoising_steps}
\tablebelow
\end{table}

\paragraph{Discussions.}
Despite the performance gains on standard benchmarks, some extremely complicated motion types, \eg, fluid dynamics, are still challenging for \ourmethod{}. Furthermore, diffusion models are computationally heavy and need distillation~\cite{Salimans:2022:PDF} for applications with a limited computational budget.

\section{Conclusion}

We have introduced a diffusion-based method for high resolution frame interpolation, named \ourmethod{}.
Our proposed \ourcascade{} cascade achieves state-of-the-art performance on several high-resolution frame interpolation benchmarks up to 8K resolution, while improving efficiency for training and inference.
We also establish a new benchmark, \ourdataset{}, which focuses on challenging cases, \eg large motion and repeated textures at high resolution. Our method substantially outperforms all methods on the benchmark as well.

\camready{
\paragraph{Acknowledgement.}
We thank Tianhao Zhang, Yisha Sun, Tristan Greszko, Christopher Farro, Fuhao Shi for their help in collecting the dataset, and Yifan Zhuang, Ming-Hsuan Yang, David Salesin, and the anonymous reviewers for their constructive feedback and discussions.
}

\clearpage 
\newpage

\appendix

\begin{center}
{\LARGE\bf Supplementary material \par}%
\end{center}

\section{Overview}
Here, we provides further implementation details, analyses on our design choices, further discussions on results on Vimeo-90K and SEPE 8K benchmark datasets, and analyses on computational complexity.
We also provide more qualitative comparison including interactive tools and videos with state-of-the-art methods in our website.
For more details, please browse our project webpage: \textit{\textbf{https://hifi-diffusion.github.io}}

\section{Implementation details}

We include further implementation details continuing from the main paper.
Our implementation is based on JAX framework.
We use a fixed random seed for reproducibility.
We use 256 TPUv5e with 16 GB memory for training.
For inference, our model runs on one A100 40GB and processes up to 8K resolution without a memory problem.
Our novel \ourmethod{} cascade enables this high-resolution processing where most of the methods have difficulties in.

\section{Effect of the patch size \camready{and overlap}}

We provide an analysis on how patch size in the cascade model affects the accuracy during inference.
Due to the self-attention layers at two bottom levels, the performance of our method could vary, when a patch size that is different from the training resolution (\ie, $224\times288$) is used.
Also, bigger patch sizes can give better accuracy due to a larger context window, but it is not so clear if it always holds.
Given our standard setup (\ie, a three-stage cascade, 4 sampling steps, an average of 4 samples), we try different patch size and evaluate on X-TEST 4K dataset~\cite{Sim:2021:XVF}.

Table~\ref{tab:supp:patch_size} reports PSNR and SSIM on X-TEST 4K dataset. 
Although the training resolution is at $224\times288$, the method is not very sensitive to the choice of patch size at inference time.
The smallest and the biggest patch size (\ie, $256\times384$ and $768\times1152$) show marginal difference in both PSNR and SSIM metrics.
The patch size $512\times768$ gives the best accuracy on X-TEST 4K.

\camready{
We also analyze the impact of varying patch overlap on X-TEST 4K, using a patch size of $512 \times 768$.
Increase of overlap size between patches can have a similar effect of averaging more samples.
By default at inference, we place patches to cover the entire image with minimal, equally distributed overlap, which is automatically determined.
In this study, we gradually increase the number of patches at each row or column with equal distanced, compute an overlap ratio, and also report PSNR X-TEST 4K benchmark. 
The overlap ratio is the value obtained by dividing the sum of all areas processed by the patches by the total image size, computed by $\frac{\text{patch size} \times \text{the number of patches}}{\text{image size}}$.

As in Table~\ref{tab:supp:patch_overlap}, overlap size does not significantly affect the performance, showing standard deviation of 0.052 for PSNR and 0.00059 for SSIM.
More overlap marginally improves the performance by averaging multiple samples but with the cost of runtime increase.
Overlap ratio with 1.33 is our default setup.
}

\begin{table}[!t]
\centering
\scriptsize
\setlength\tabcolsep{4.5pt}
\caption{\textbf{Effect of different patch size:} The usage of different patch sizes does not show significant accuracy difference on X-TEST 4K dataset.}
\begin{tabularx}{0.5\linewidth}{@{} X @{\hskip 2em} *{2}{S[table-format=2.3]}}
\toprule
Patch size & PSNR & SSIM \\ \midrule
{$256\times384$} & 32.70 & \secondbest 0.931 \\
{$384\times576$} & \secondbest 32.82 & \best 0.932 \\
{$512\times768$} & \best 32.92 & \secondbest 0.931 \\
{$640\times960$} & 32.79 & 0.930 \\
{$768\times1152$} & 32.75 & 0.930 \\
\bottomrule
\end{tabularx}
\label{tab:supp:patch_size}
\end{table}

\begin{table}[!t]
\centering
\scriptsize
\setlength\tabcolsep{3pt}
\caption{\textbf{\camready{Effect of overlapping ratio between patches:}} Overlap size marginally affect the accuracy. The overlap ratio with 1.33 is our default setup.}
\begin{tabularx}{\linewidth}{@{} X @{\hskip 1em} *{8}{c}}
\toprule
Overlap ratio & 1.33 & 1.56 & 1.60 & 1.78 & 1.87 & 2.13 & 2.18 & 2.49 \\
\midrule
PSNR & 32.92 & 32.99 & 33.00 & 33.02 & 32.98 & 33.10 & 33.03 & 33.07 \\
SSIM & 0.931 & 0.932 & 0.932 & 0.932 & 0.932 & 0.933 & 0.932 & 0.933 \\
\bottomrule
\end{tabularx}
\label{tab:supp:patch_overlap}
\end{table}

\section{Dropout for \ourcascade{} cascade training}
\label{sec:supp:dropout_effect}

\begin{figure}[!t]
\newcommand{\dropoutimgwidth}{0.24\linewidth}
\centering
\footnotesize
\newcommand{\datasetimgwidth}{0.24\linewidth}
\newcommand{\colskip}{0.1em}
\begin{tabular}{c @{\hskip \colskip} c @{\hskip \colskip}  c @{\hskip \colskip} c}
\includegraphics[width=\datasetimgwidth]{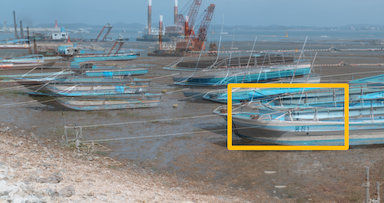} & 
\includegraphics[width=\datasetimgwidth]{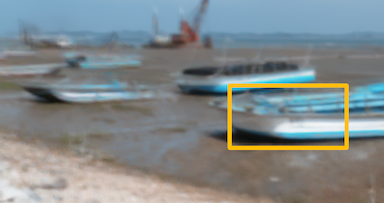} & 	\includegraphics[width=\datasetimgwidth]{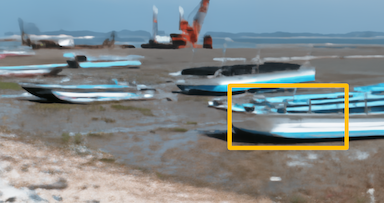}& 
\includegraphics[width=\datasetimgwidth]{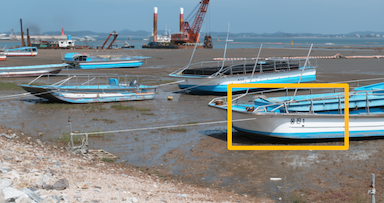} \\
\includegraphics[width=\datasetimgwidth]{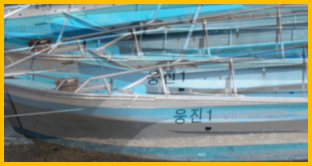} & 
\includegraphics[width=\datasetimgwidth]{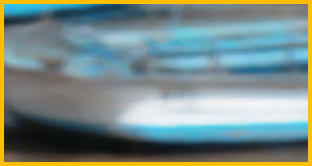} & 
\includegraphics[width=\datasetimgwidth]{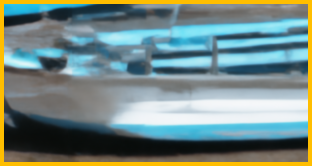}& 
\includegraphics[width=\datasetimgwidth]{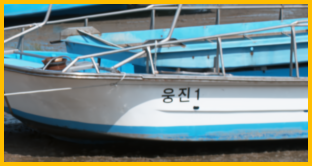} \\
Overlaid inputs & Intermediate & W/o dropout & W/ dropout
\end{tabular}
\caption{\textbf{Dropout} forces the network to use both the low resolution intermediate and the inputs. Without dropout, the network takes a shortcut and only tries to upsamples the intermediate with missing details. The second row shows close views of highlighted areas in the images at the first row.
}
\label{fig:ablation_dropout}
\end{figure}

\begin{figure*}[!ht]
\centering
\definecolor{dropoutcellgreen}{RGB}{219,242,224}
\definecolor{dropoutcellblue}{RGB}{228,247,255}
\begin{minipage}{0.20\textwidth}
\subcaptionbox{Input overlay \label{fig:supp:dropout_effect:input}}{
\includegraphics[width=\textwidth]{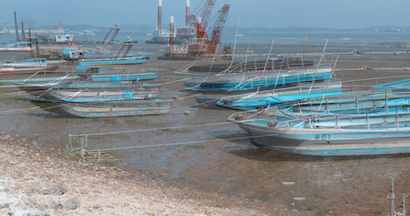}}
\subcaptionbox{Ground truth\label{fig:supp:dropout_effect:gt}}{
\includegraphics[width=\textwidth]{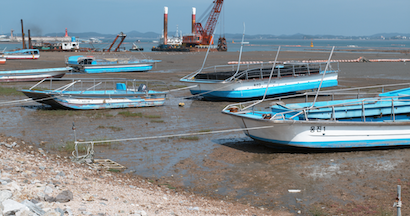}}
\end{minipage} \hspace{3em}
\begin{minipage}{0.48\textwidth}
\subcaptionbox{Model diagram\label{fig:supp:dropout_effect:arch}}{\includegraphics[width=\textwidth]{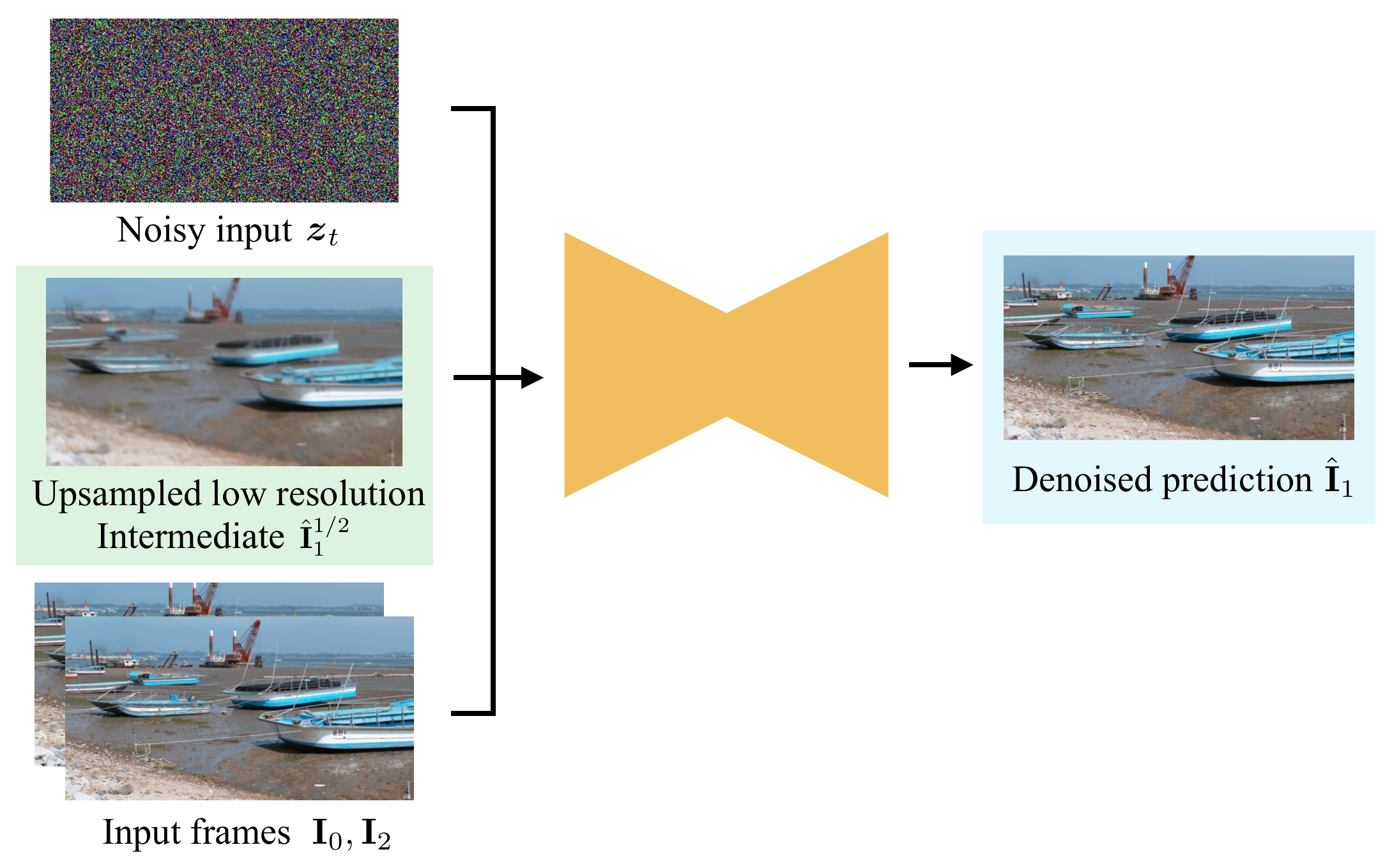}}
\end{minipage}
\\[0.5em]
\newcommand{\datasetimgwidth}{0.226\linewidth}
\subcaptionbox{Model with dropout\label{fig:supp:dropout_effect:w_dropout}}{
\centering
\footnotesize
\setlength\tabcolsep{0.4pt}
\begin{tabular}{c @{\hskip 0.3em} c @{\hskip 0.3em} c}
    & Oracle & Different \\
    \rotatebox{90}{\hspace{0.8em}\makecell{\cellcolor{dropoutcellgreen}Low-res\\\cellcolor{dropoutcellgreen}intermediate}} &
	\includegraphics[width=\datasetimgwidth]{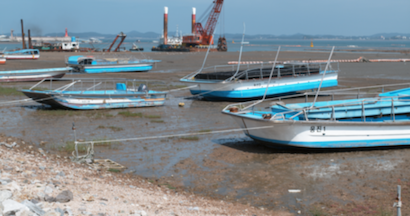} & 
	\includegraphics[width=\datasetimgwidth]{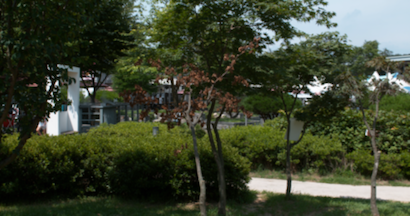} \\
	\rotatebox{90}{\hspace{1.2em}\makecell{\cellcolor{dropoutcellblue}Denoised\\\cellcolor{dropoutcellblue}prediction}} &
	\includegraphics[width=\datasetimgwidth]{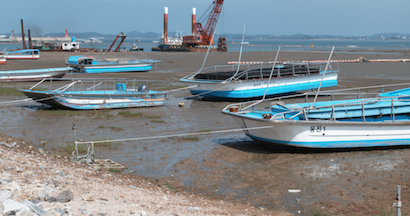} & 
	\includegraphics[width=\datasetimgwidth]{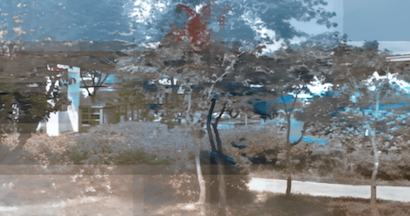} \\
	\rotatebox{90}{\hspace{1em}\makecell{Error map}} &
	\includegraphics[width=\datasetimgwidth]{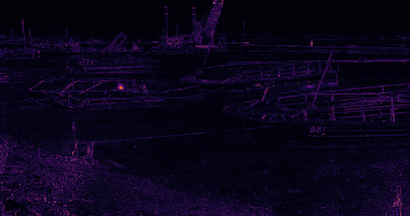} & 
	\includegraphics[width=\datasetimgwidth]{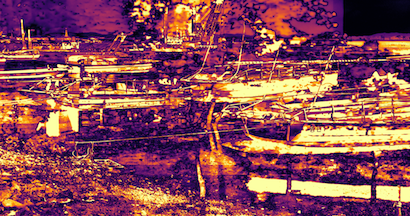}
\end{tabular}
}
\subcaptionbox{Model without dropout\label{fig:supp:dropout_effect:wo_dropout}}{
\centering
\footnotesize
\setlength\tabcolsep{0.4pt}
\begin{tabular}{c @{\hskip 0.3em} c}
    Oracle & Different \\
	\includegraphics[width=\datasetimgwidth]{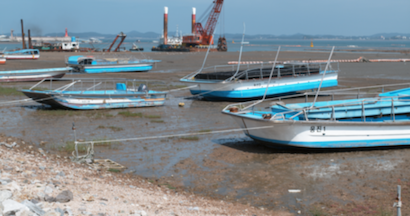} & 
	\includegraphics[width=\datasetimgwidth]{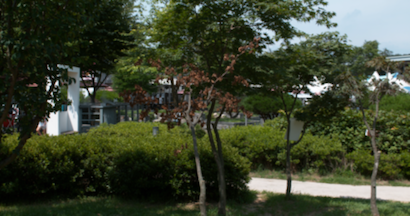} \\
	\includegraphics[width=\datasetimgwidth]{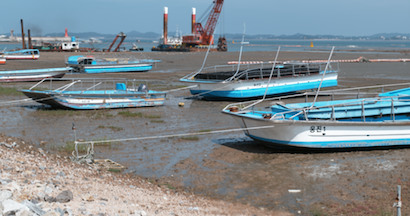} & 
	\includegraphics[width=\datasetimgwidth]{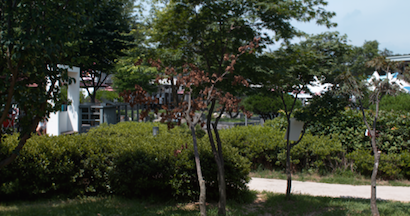} \\ 
	\includegraphics[width=\datasetimgwidth]{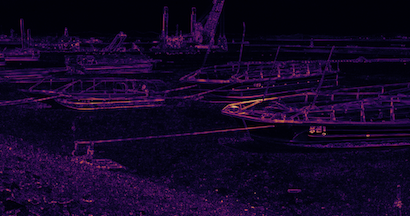} & 
	\includegraphics[width=\datasetimgwidth]{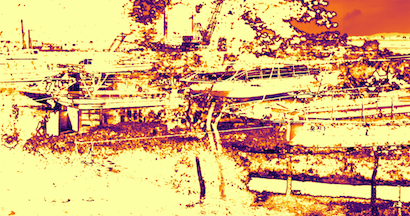}
\end{tabular}
}
\caption{
\textbf{Effect of dropout}. In the two-stage cascade formulation, we train our cascade model with or without dropout and visualize results when inputting oracle/different low resolution intermediate respectively.
These inputs are downsampled and upsampled back to the original resolution to mimic the low resolution intermediate from the previous level.
\emph{(d)} With dropout the model effectively utilizes coarse structure from the intermediate and fine details from the high resolution input.
This holds true even with different low resolution intermediate: the model add color and texture to the coarse structure.
\emph{(e)} On the other hand, the model without dropout solely relies on the intermediate, primarily sharpening it.
This leads to a loss of fine details, \eg around object boundaries (see the error map).
The behavior becomes looking more apparent with a different intermediate; the model ignores input frames.
}
\label{fig:supp:dropout_effect}
\end{figure*}

We find that the image-level dropout is crucial for making the \ourcascade{} cascade model behave as intended, especially for challenging large motion scenes, as in \cref{fig:ablation_dropout}.
Without dropout, the model finds a shortcut, sharpening the low resolution intermediate from the base model without referring to input images. This results in losing fine details, \eg, thin structures and letters.
With dropout, the model refers to both the low resolution intermediate for coarse structure and the high-resolution input for fine details.
In this study, we train the model on X-TRAIN~\cite{Sim:2021:XVF} only, as the network without dropout does not converge with a full training dataset.
This suggests that the dropout also stabilizes large scale training for the cascade model.

To further analyze the effect of the image-level dropout, we prepare two models that are with or without the dropout and see the models' behavior by inputting a different image as the low resolution intermediate during the inference.
We test a two-stage cascade model as in \cref{fig:supp:dropout_effect:arch}.

The model with dropout (\ie, \cref{fig:supp:dropout_effect:w_dropout}) successfully outputs high-resolution prediction when actual low resolution intermediates are inputted. 
When a different intermediate is inputted, the model tries to add appearance (\eg, color or texture) from input frames on top of object structures from the low resolution intermediate.
Though it produces non-sensible output, this proves that the model is able to exploit both input frame and low resolution intermediate during the inference.

On the other hand, the model trained without dropout (\ie,~\cref{fig:supp:dropout_effect:wo_dropout}) only refers to the low resolution intermediate. Even when inputting a different image as the low resolution condition (\eg,~the right column in~\cref{fig:supp:dropout_effect:wo_dropout}), the denoised prediction completely ignores the input frames and takes a shortcut to sharpen the low resolution condition (\ie, tree image). 
Our probabilistic image-level dropout prevents the model from taking this shortcut and learns to refer to both input and condition cues.

\section{Effect of the number of sampling steps}

\begin{figure*}[!ht]
\centering
\footnotesize
\setlength\tabcolsep{0.75pt}

\subcaptionbox{Two input frames (above) and ground truth (below)\label{fig:supp:vimeo_samples_input}}{
\begin{tabular}{cc}
    \includegraphics[width=0.22\textwidth]{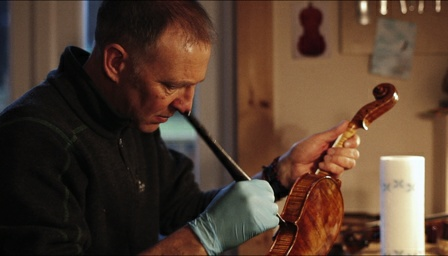} &
    \includegraphics[width=0.22\textwidth]{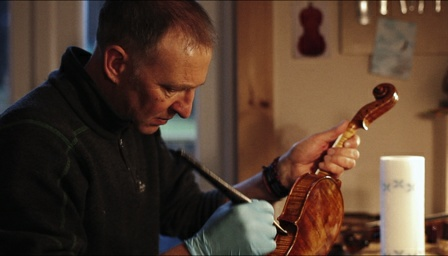} \\
    \multicolumn{2}{c}{\includegraphics[width=0.22\textwidth]{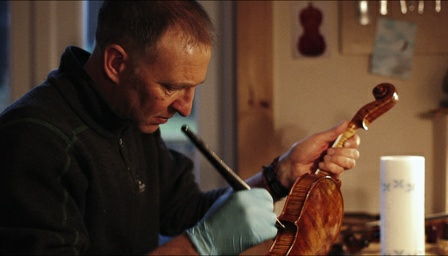}} \\
\end{tabular}
} \hspace{2em}
\subcaptionbox{Multiple samples with 64 sampling steps\label{fig:supp:vimeo_samples_64_steps}}{
\begin{tabular}{cc}
    \includegraphics[width=0.22\textwidth]{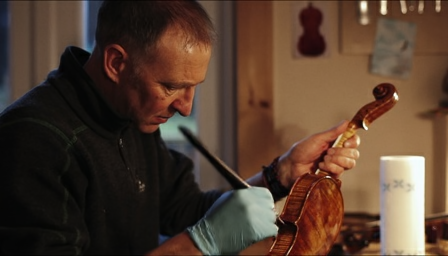} &
    \includegraphics[width=0.22\textwidth]{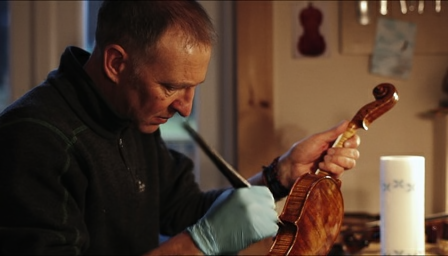} \\
    \includegraphics[width=0.22\textwidth]{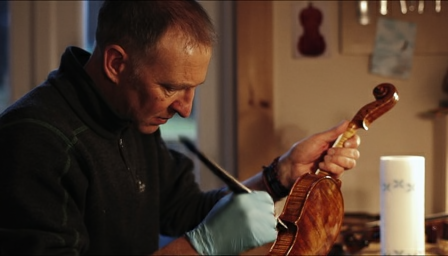} &
    \includegraphics[width=0.22\textwidth]{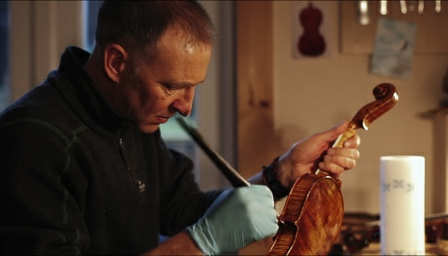} \\
\end{tabular}
} \\[2em]
\subcaptionbox{Average sample, error map, and variance map \wrt the different number of sampling steps. \label{fig:supp:vimeo_samples_steps}}{
\newcommand{\vimeosamplewidth}{0.17\textwidth}
\newcommand{\vimeosamplerow}[1]{
\raisebox{-.5\height}{\includegraphics[width=\vimeosamplewidth]{figures/vimeo/samples/1_#1.png}} &
\raisebox{-.5\height}{\includegraphics[width=\vimeosamplewidth]{figures/vimeo/samples/4_#1.png}} &
\raisebox{-.5\height}{\includegraphics[width=\vimeosamplewidth]{figures/vimeo/samples/8_#1.png}} &
\raisebox{-.5\height}{\includegraphics[width=\vimeosamplewidth]{figures/vimeo/samples/16_#1.png}} &
\raisebox{-.5\height}{\includegraphics[width=\vimeosamplewidth]{figures/vimeo/samples/64_#1.png}}
}
\begin{tabular}{c @{\hskip 1em} c *{4}{@{\hskip 0.3em} c}}
    \makecell{Average\\sample} & \vimeosamplerow{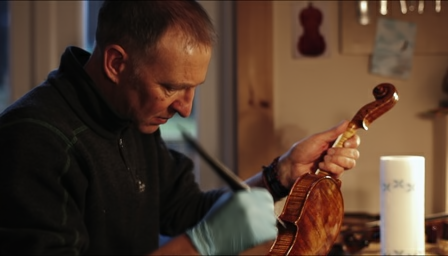} \\ 
    \makecell{Error\\map} & \vimeosamplerow{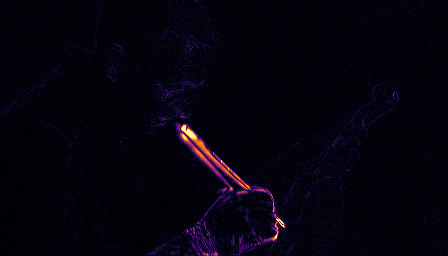} \\ 
    \makecell{Variance\\map} & \vimeosamplerow{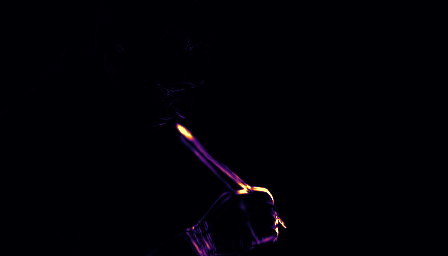} \\
    \makecell{Sampling\\steps} & 1 & 4 & 8 & 16 & 64 \\
\end{tabular}
} \\[1em]

\caption{\textbf{Effect of the number of sampling steps}: We visualize how the number of sampling steps affects results. (a) Given two input frames, we try different sampling steps and visualize (b) each individual sample as well as (c) averaged sample, error map, and variance map. With more sampling steps, the model predicts multiple plausible diverse samples with non-linear motion (\ie, the fast moving stick).}
\label{fig:supp:vimeo_samples}
\end{figure*}

In \cref{fig:supp:vimeo_samples}, we visualize how the number of sampling steps affects results.
The input frame in~\cref{fig:supp:vimeo_samples_input} shows a person using a stick, and the stick moves fast between the frames. 
The variance map in \cref{fig:supp:vimeo_samples_steps} shows that with more sampling steps, the model outputs more diverse motion of the stick and the hand of the person, as highlighted in the various map.
With the lower number of sampling steps, our model produces close-to-mean prediction.
\cref{fig:supp:vimeo_samples_64_steps} visualizes the four samples drawn from 64 sampling steps; our model predicts diverse, plausible samples with non-linear motion, such as in different trajectories or at different acceleration and deceleration rate.
This follows the same observation from DDVM~\cite{Saxena:2023:TSE} that the diffusion model is able to predict plausible multi-mode samples. 

\camready{
\section{Randomness and robustness}
With the stochastic nature, \ourmethod{} predicts plausible diverse samples with non-linear motion as in \cref{fig:supp:vimeo_samples}, as a unique capability. 
To see if this unique property can also affect accuracy, we tested 10 runs with different random seeds and compute mean and standard deviation of results on four difference benchmark datasets.
As reported in \cref{tab:supp:rand_robust}, it does not yield much variation on accuracy.
}
\begin{table}[!t]
\centering
\scriptsize
\setlength\tabcolsep{4.5pt}
\caption{\camready{\textbf{Randomness and robustness}: the unique tochastic nature from diffusion does not yield much variation on accuracy.}}
\begin{tabularx}{0.8\linewidth}{@{} X @{\hskip 2em} *{4}{c}}
\toprule
PSNR & Vimeo & Xiph 2K & X-TEST 4K & LaMoR \\
\midrule
mean & 36.12 & 37.34 & 32.92 & 28.15 \\
std.~dev. & 0.0052 & 0.0024 & 0.0196 & 0.0233 \\
\bottomrule
\end{tabularx}
\label{tab:supp:rand_robust}
\end{table}

\begin{table}[!t]
\centering
\scriptsize
\setlength\tabcolsep{4.5pt}
\caption{\textbf{Results on SEPE 8K dataset.} \ourmethod{} outperforms  M2M~\cite{Hu:2022:M2M}. Most of the methods have the OOM (out of memory) problem at 8K resolution.}
\begin{tabularx}{0.8\linewidth}{@{} X @{\hskip 2em} *{2}{S[table-format=2.3]}}
\toprule
Method & PSNR & SSIM\\\midrule
{LDMVFI~\cite{Danier:2024:LDM}} & \multicolumn{2}{c}{OOM} \\
{EMA-VFI~\cite{Zhang:2023:EMA}} & \multicolumn{2}{c}{OOM} \\
{UPR-Net~\cite{Jin:2023:UPR}} & \multicolumn{2}{c}{OOM} \\
{BiFormer~\cite{Park:2023:BFL}} & \multicolumn{2}{c}{OOM} \\
{M2M~\cite{Hu:2022:M2M}} & 28.34 & \secondbest 0.883 \\
{SGM-VFI~\cite{Liu_2024_CVPR}} & \secondbest 28.43 & 0.880 \\
{\textbf{\ourmethod{} (Ours)}} & \best 29.78 & \best 0.900 \\
\bottomrule
\end{tabularx}
\label{tab:benchmark_sepe8k}
\end{table}

\begin{figure*}[!ht]
\centering
\footnotesize
\setlength\tabcolsep{1pt}

\newcommand{\sepeimgoverlay}[1]{\includegraphics[width=0.36\textwidth]{figures/qual/sepe8k/#1_overlay.png}}
\newcommand{\sepeimg}[3]{\raisebox{-.85\height}{\includegraphics[width=0.15\textwidth]{figures/qual/sepe8k/#1_#2_#3.png}}}
\newcommand{\seperow}[1]{
\multirow{2}{*}{\sepeimgoverlay{#1}} & \sepeimg{#1}{m2m}{1} & \sepeimg{#1}{sgm}{1} & \sepeimg{#1}{ours}{1} & \sepeimg{#1}{gt}{1} \\[4.3em]
& \sepeimg{#1}{m2m}{2} & \sepeimg{#1}{sgm}{2} & \sepeimg{#1}{ours}{2} & \sepeimg{#1}{gt}{2}}
\begin{tabular}{ccccc}
    \seperow{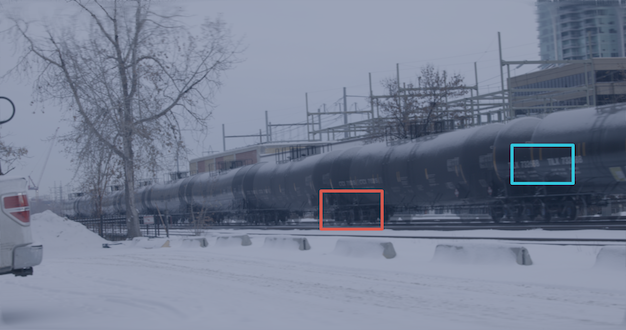} \\[5em]
    \seperow{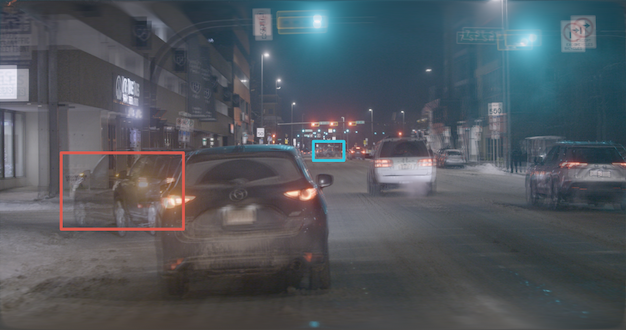} \\[5em] 
    \seperow{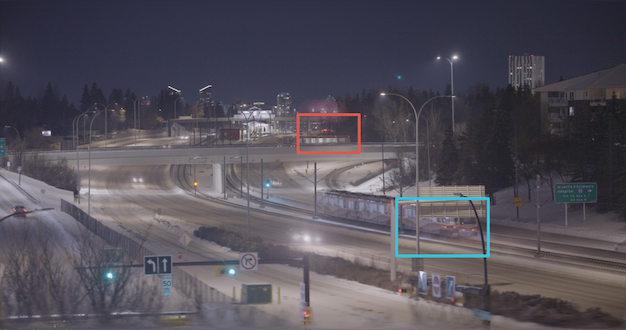} \\[5em] 
    \seperow{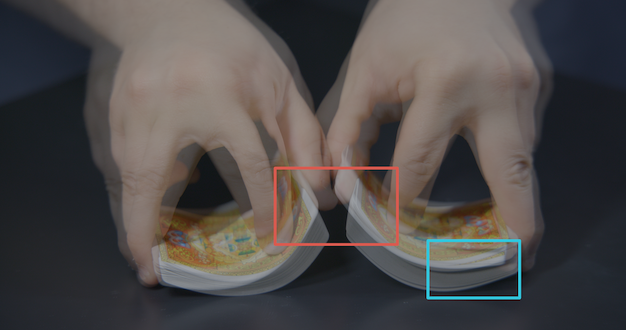} \\[5em]
    Input overlay & M2M & SGM-VFI & \textbf{\ourmethod{} (Ours)} & Ground Truth \\
\end{tabular}
\caption{\textbf{Qualitative comparison on SEPE 8K}: We provide qualitative comparison between our method and M2M~\cite{Hu:2022:M2M} which is able to run at 8K resolution. 
Compared to M2M~\cite{Hu:2022:M2M} and SGM-VFI~\cite{Liu_2024_CVPR} which have a difficulty in handling large motion, our method is able to recover fine details on challenging cases at 8K resolution.}
\label{fig:supp:m2m_comparison}
\end{figure*}

\section{Evaluation on SEPE 8K benchmark}

SEPE 8K dataset~\cite{al2023sepe} provides 40 raw videos with 300 frames, 8K resolution, and 29.97 FPS for benchmarking various downstream computer vision tasks such as video quality assessment, super-resolution, compression, \etc.
To utilize the dataset for benchmarking frame interpolation methods, especially for high resolution with large motion, we select a triple of frames (145th, 150th, and 155th) from each video and target to predict the middle frame from the rest two frames as input.

Table~\ref{tab:benchmark_sepe8k} includes all state-of-the-art methods that we compare on SEPE 8K benchmark. 
Except M2M~\cite{Hu:2022:M2M} and SGM-VFI~\cite{Liu_2024_CVPR}, the other methods are not able to process 8K resolution image due to the out-of-memory (OOM) problem on A100 40GB GPU.
Figure~\ref{fig:supp:m2m_comparison} provide qualitative comparison between our method and M2M~\cite{Hu:2022:M2M}.
Unlike M2M~\cite{Hu:2022:M2M}, our method is able to recover fine details on such challenging large motion cases even at 8K resolution.

\begin{figure*}[!ht]
\centering
\footnotesize
\setlength\tabcolsep{0.75pt}

\newcommand{\vimeoerrimg}[2]{\includegraphics[width=0.24\textwidth]{figures/vimeo/error_map/#1_#2.png}}
\newcommand{\vimeoerrrow}[1]{
\vimeoerrimg{#1}{overlay} & \vimeoerrimg{#1}{pred} & \vimeoerrimg{#1}{tgt} & \vimeoerrimg{#1}{err}
}

\begin{tabular}{cccc}
    \vimeoerrrow{227} \\ 
    \vimeoerrrow{277} \\ 
    \vimeoerrrow{331} \\ 
    \vimeoerrrow{6} \\ 
    \vimeoerrrow{846} \\ 
    \vimeoerrrow{963} \\ 
    \vimeoerrrow{988} \\ 
    Input overlay & Ours & Ground truth & Error map \\
\end{tabular}
\caption{\textbf{Error map visualization on Vimeo-90K}: We randomly sample ours results with high errors on Vimeo-90k and visualize them with input overlay, ground truth, and error map. The error mostly originates from motion boundaries where the predicted objects' motion sometimes do not align well with true motion. This is because Vimeo-90K dataset prefers linear motion prediction whereas our model can predict plausible non-linear motion~\cite{Kiefhaber:2024:BVF}.}
\label{fig:supp:vimeo_error}
\end{figure*}

\section{Discussion on Vimeo-90K}

While \ourmethod{} achieves the best accuracy on multiple high resolution benchmark datasets, it performs comparably on highly-saturated Vimeo-90K benchmark~\cite{Xue:2019:VEW}.
To analyze the behavior, in~\cref{fig:supp:vimeo_error} we show random-sampled results of our method that gives high errors on Vimeo-90K.
As visualized in the error map, erroneous prediction mostly arises from motion boundaries even if \ourmethod{} is able to interpolate frames with fine details, as shown in the second column.
This is specifically due to the dataset property: Vimeo-90K prefers linear motion prediction~\cite{Kiefhaber:2024:BVF} whereas our model predicts diverse non-linear motion of objects.
Non-linear motion prediction yields a subtle misalignment between predicted position and true position of object, and it causes intensity difference mostly near image edges.
Thus errors mostly arise near object or motion boundaries.
In \cref{fig:supp:vimeo_samples_64_steps}, we visualize multiple non-linear motion examples that our model produces.

\begin{figure*}[!t]
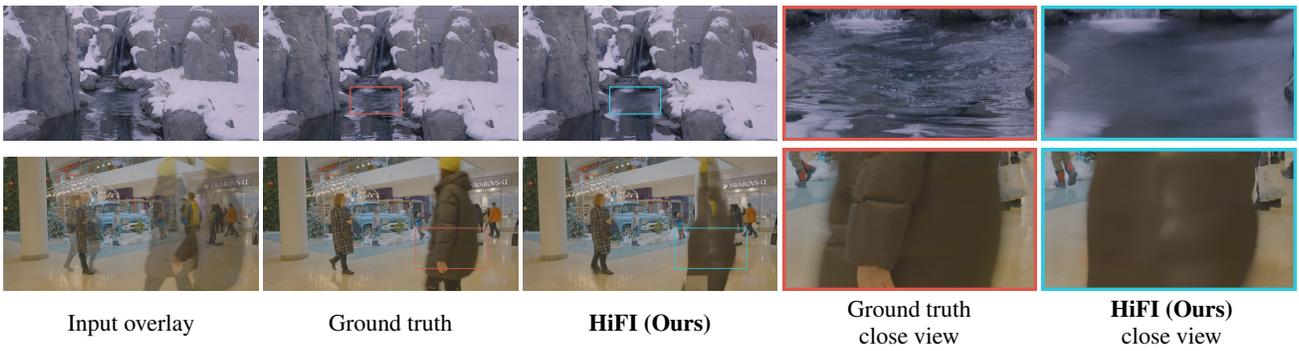

\centering
\footnotesize
\setlength\tabcolsep{1pt}

\newcommand{\limitationimage}[2]{\includegraphics[width=0.19\textwidth]{figures/limitation/#1_#2.png}}
\newcommand{\limitationrow}[1]{
\limitationimage{#1}{overlay} & \limitationimage{#1}{gt} & \limitationimage{#1}{ours} & \limitationimage{#1}{gt_crop} & \limitationimage{#1}{ours_crop}
}
\begin{tabular}{ccccc}
    \limitationrow{sepe} \\
    \limitationrow{sepe2} \\
    Input overlay & Ground truth & \textbf{\ourmethod{} (Ours)} & \makecell{Ground truth\\close view} & \makecell{\textbf{\ourmethod{} (Ours)}\\close view} \\
\end{tabular}
\caption{\textbf{Some extremely challenging cases} that even our model struggles with: \emph{(above)} fluid motion where more generative solution can be preferred or \emph{(below)} extremely large motion, around 1500 pixels in the example, that is much bigger than the patch size at inference time.}
\label{fig:supp:limitation}
\end{figure*}

\camready{
\section{Comparison with diffusion-based approach}

Between two existing diffusion-based methods, VIDIM~\cite{Jain:2024:VID} and LDMVFI~\cite{Danier:2024:LDM}, we provide comparisons with LDMVFI~\cite{Danier:2024:LDM} in \cref{tab:supp:comparison_diffusion}.
VIDIM~\cite{Jain:2024:VID} focuses on a task closer to conditional video generation, a long-range interpolation on 256 $\times$ 256 resolution, which is different from our target task.

Our method consistently outperforms LDMVFI \cite{Danier:2024:LDM} by a large margin, even on LPIPS metric despite that LDMVFI \cite{Danier:2024:LDM} uses LPIPS-based training loss.
Furthermore, LDMVFI \cite{Danier:2024:LDM} does not run on SEPE 8K due to the out of memory problem.
}

\begin{table}[!t]
\centering
\scriptsize
\setlength\tabcolsep{5pt}
\caption{\textbf{\camready{Comparison with diffusion-based approaches:}} Our method consistently outperforms LDMVFI~\cite{Danier:2024:LDM} on both X-TEST 4K and LaMoR datasets by a large margin.}
\begin{tabularx}{\linewidth}{@{} X @{\hskip 1em} *{6}{c}}
\toprule
& \multicolumn{3}{c}{X-TEST 4K} & \multicolumn{3}{c}{LaMoR} \\
\cmidrule(lr){2-4} \cmidrule(lr){5-7}
& PSNR $\uparrow$ & SSIM $\uparrow$ & LPIPS $\downarrow$ & PSNR $\uparrow$ & SSIM $\uparrow$ & LPIPS $\downarrow$\\
\midrule
\textbf{Ours} & \best32.92 & \best0.931 & \best0.2136 &\best 28.141 & \best0.912 &\best 0.1880 \\
LDMVFI & 23.36 & 0.770 & 0.3285 & 21.952 & 0.828 & 0.2438 \\
\bottomrule
\end{tabularx}
\label{tab:supp:comparison_diffusion}
\end{table}

\section{\camready{Computational complexity}}
\camready{
\cref{tab:supp:computation} provides comparison on computational complexity and runtime performance among different methods.
We test each method on X-TEST 4K benchmark and report its accuracy, inference time, and peak memory, using one A100 40GB GPU.
Despite of its demanding model size and runtime, our method requires only 7.53 GB peak memory for 4K image processing and achieves the best accuracy compared with others.

We also provide analysis on the impact of cascade levels (\cref{tab:supp:computation_cascade}) and patch size (\cref{tab:supp:computation_patch}) on accuracy and runtime (X-TEST 4K, A100 40GB).
As in \cref{tab:supp:computation_cascade}, the direct input of 4K images causes out-of-memory errors. Simple adoption of patch-based processing (\ie (Patch-based) Base) reduces memory, though it struggles with large motion, resulting in lower PSNR and SSIM.
Our patch-based cascades eventually handle challenging cases with improved accuracy, with peak-memory nearly unchanged.
\cref{tab:supp:computation_patch} provide an analysis that both peak memory and runtime increase \wrt the patch size mainly due to the self-attention layers in the bottleneck.
We find that the patch size of 512 $\times$ 768 achieves the best performance without having too much increase of runtime and memory.

We leave the reduce of computational cost as future work. 
Runtime can be substantially reduced by using fewer denoising steps.
Having just two steps can yield up to 2 times speedup with only a 0.16\% average accuracy drop across five benchmark datasets.
Batch processing of samples and patches will further accelerate the runtime while maintaining the same accuracy.
Model parameters can also be reduced via model distillation.
}

\begin{table}[!t]
\centering
\scriptsize
\caption{\textbf{\camready{Comparison on computational complexity:}} We measure computational complexity and accuracy of each model on X-TEST 4K benchmark. Our method requires only 7.53 GB peak memory for 4K image processing and achieves the best accuracy compared with others.}
\begin{tabularx}{\linewidth}{@{} X @{\hskip 2em} *{4}{S[table-format=2.3]}}
\toprule
Method & {\makecell{Inference\\time (s)}} & {\makecell{Model\\parameters (M)}} & {\makecell{Peak\\memory (GB)}} & {\makecell{X-TEST 4K\\PSNR (dB)}} \\ \midrule
UPR-Net & \best 1.96 & \best6.56 & 33.44 & 30.68 \\
M2M &  2.51 & \secondbest7.61 & \secondbest12.65 & 30.81 \\
BiFormer & \secondbest 2.34 & 11.17 & 21.50 & 31.32 \\
EMA-VFI & 3.27 & 65.66 & 30.30 & \secondbest31.46 \\
LDMVFI & 11.63 & 416.46 & 37.23 & 23.36 \\
\textbf{HiFI (ours)} & 164.18 & 647.74 & \best7.53 & \best32.92 \\
\bottomrule
\end{tabularx}
\label{tab:supp:computation}
\end{table}

\begin{table}[!t]
\centering
\scriptsize
\caption{\textbf{\camready{Efficiency analysis of cascade processing:}} Our patch-based cascades handle challenging cases with improved accuracy, with peak-memory nearly unchanged.
}
\begin{tabularx}{\linewidth}{@{} X @{\hskip 2em} *{4}{c}}
\toprule
Method & {\makecell{Peak\\memory (GB)}} & {\makecell{runtime (s)}} & {PSNR} & {SSIM} \\ \midrule
Whole-image processing & OOM & -- & -- & -- \\ \midrule
(Patch-based) Base & 7.47 & 94.62 & 28.57 & 0.876 \\
Patch-based cascade $\times$ 2 & 7.52 & 136.53 & 32.77 & 0.930 \\
\textbf{Patch-based cascade $\times$ 3} & 7.53 & 164.18 & 32.92 & 0.931 \\
\bottomrule
\end{tabularx}
\label{tab:supp:computation_cascade}
\end{table}

\begin{table}[!t]
\centering
\scriptsize
\setlength\tabcolsep{2pt}
\caption{\textbf{\camready{Efficiency analysis of patch-wise processing:}} 
The usage of the patch size of 512 $\times$ 768 achieves the best performance with a reasonable increase of runtime and memory.
}
\begin{tabularx}{\linewidth}{@{} X l @{\hskip 3.5em} *{4}{c}}
\toprule
 & Patch size & {\makecell{Peak\\memory (GB)}} & Runtime (s) & PNSR & SSIM \\
 \midrule
\multirow{4}{*}{Patch-based} & 384$\times$576 & 6.49 & 139.68 & 32.82 & 0.932 \\
 & \textbf{512$\times$768} & 7.53 & 164.18 & 32.92 & 0.931 \\
 & 640$\times$960 & 9.34 & 189.57 & 32.79 & 0.930 \\
 & 768$\times$1152 & 12.51 & 209.80 & 32.75 & 0.930 \\  \midrule
Whole image & 2160$\times$4096 & OOM  & -- & -- & -- \\
\bottomrule
\end{tabularx}
\label{tab:supp:computation_patch}
\end{table}

\section{Challenges and future work}

Our method achieves the state of the art on multiple high-resolution benchmark datasets, yet there still exists some extremely difficult cases that challenge our method.
Figure~\ref{fig:supp:limitation} shows a few examples from the SEPE 8K dataset.
In case of fluid motion in the first row, \ourmethod{} tends to output blurry results; more generative solution can be preferred. 
Also as in the second row, when motion is extremely larger (\eg 1500 pixels) than the patch size, most of the content goes out of image boundary. \ourmethod{} cannot establish reliable correspondence, and thus is not able to interpolate their motion.
\camready{The usage of a bigger patch size can resolve the issues but with increased computational cost.}

\camready{
Furthermore, our model predicts one middle frame given two input frames as a basic setup, following the conventional setup.
As future work, our model can be easily extended to the multi-frame setup by conditioning on multiple frames and interpolating multiple middle frames together, trained on raw videos. This multi-frame setup can also effectively represent non-linear motion.
}

\bibliography{main}

\begin{thebibliography}{59}
\providecommand{\natexlab}[1]{#1}

\bibitem[{Al~Shoura et~al.(2023)Al~Shoura, Dehaghi, Razavi, Far, and
  Moshirpour}]{al2023sepe}
Al~Shoura, T.; Dehaghi, A.~M.; Razavi, R.; Far, B.; and Moshirpour, M. 2023.
\newblock {SEPE} {D}ataset: {8K} Video Sequences and Images for Analysis and
  Development.
\newblock In \emph{Conference on ACM Multimedia Systems}, 463--468.

\bibitem[{Bar-Tal et~al.(2023)Bar-Tal, Yariv, Lipman, and
  Dekel}]{bar2023multidiffusion}
Bar-Tal, O.; Yariv, L.; Lipman, Y.; and Dekel, T. 2023.
\newblock Multi{D}iffusion: Fusing diffusion paths for controlled image
  generation.
\newblock In \emph{ICML}.

\bibitem[{Baranchuk et~al.(2022)Baranchuk, Voynov, Rubachev, Khrulkov, and
  Babenko}]{Baranchuk:2022:LES}
Baranchuk, D.; Voynov, A.; Rubachev, I.; Khrulkov, V.; and Babenko, A. 2022.
\newblock Label-Efficient Semantic Segmentation with Diffusion Models.
\newblock In \emph{ICLR}.

\bibitem[{Blattmann et~al.(2023)Blattmann, Dockhorn, Kulal, Mendelevitch,
  Kilian, Lorenz, Levi, English, Voleti, Letts, Jampani, and
  Rombach}]{Blattmann:2023:SVD}
Blattmann, A.; Dockhorn, T.; Kulal, S.; Mendelevitch, D.; Kilian, M.; Lorenz,
  D.; Levi, Y.; English, Z.; Voleti, V.; Letts, A.; Jampani, V.; and Rombach,
  R. 2023.
\newblock Stable video diffusion: Scaling latent video diffusion models to
  large datasets.
\newblock \emph{arXiv:2311.15127 [cs.CV]}.

\bibitem[{Brooks, Holynski, and Efros(2023)}]{Brooks:2023:IPP}
Brooks, T.; Holynski, A.; and Efros, A.~A. 2023.
\newblock InstructPix2Pix: Learning to follow image editing instructions.
\newblock In \emph{CVPR}, 18392--18402.

\bibitem[{Cheng and Chen(2020)}]{Cheng:2020:VFI}
Cheng, X.; and Chen, Z. 2020.
\newblock Video frame interpolation via deformable separable convolution.
\newblock In \emph{AAAI}, 10607--10614.

\bibitem[{Choi et~al.(2020)Choi, Kim, Han, Xu, and Lee}]{Choi:2020:CAI}
Choi, M.; Kim, H.; Han, B.; Xu, N.; and Lee, K.~M. 2020.
\newblock Channel attention is all you need for video frame interpolation.
\newblock In \emph{AAAI}, 10663--10671.

\bibitem[{Danier, Zhang, and Bull(2024)}]{Danier:2024:LDM}
Danier, D.; Zhang, F.; and Bull, D. 2024.
\newblock {LDMVFI}: Video frame interpolation with latent diffusion models.
\newblock In \emph{AAAI}, 1472--1480.

\bibitem[{Ding et~al.(2024)Ding, Zhang, Wu, and Tu}]{ding2023patched}
Ding, Z.; Zhang, M.; Wu, J.; and Tu, Z. 2024.
\newblock Patched denoising diffusion models for high-resolution image
  synthesis.
\newblock In \emph{ICLR}.

\bibitem[{Ho et~al.(2022{\natexlab{a}})Ho, Saharia, Chan, Fleet, Norouzi, and
  Salimans}]{Ho:2022:CDM}
Ho, J.; Saharia, C.; Chan, W.; Fleet, D.~J.; Norouzi, M.; and Salimans, T.
  2022{\natexlab{a}}.
\newblock Cascaded diffusion models for high fidelity image generation.
\newblock \emph{JMLR}, 23(47): 1--33.

\bibitem[{Ho and Salimans(2022)}]{ho2022classifier}
Ho, J.; and Salimans, T. 2022.
\newblock Classifier-free diffusion guidance.
\newblock \emph{arXiv preprint arXiv:2207.12598}.

\bibitem[{Ho et~al.(2022{\natexlab{b}})Ho, Salimans, Gritsenko, Chan, Norouzi,
  and Fleet}]{Ho:2022:VDM}
Ho, J.; Salimans, T.; Gritsenko, A.; Chan, W.; Norouzi, M.; and Fleet, D.~J.
  2022{\natexlab{b}}.
\newblock Video diffusion models.
\newblock \emph{NeurIPS}, 35: 8633--8646.

\bibitem[{Hu et~al.(2024)Hu, Jiang, Zhong, Wang, and Zheng}]{Hu_2024_CVPR}
Hu, M.; Jiang, K.; Zhong, Z.; Wang, Z.; and Zheng, Y. 2024.
\newblock {IQ-VFI}: Implicit Quadratic Motion Estimation for Video Frame
  Interpolation.
\newblock In \emph{CVPR}, 6410--6419.

\bibitem[{Hu et~al.(2022)Hu, Niklaus, Sclaroff, and Saenko}]{Hu:2022:M2M}
Hu, P.; Niklaus, S.; Sclaroff, S.; and Saenko, K. 2022.
\newblock Many-to-many splatting for efficient video frame interpolation.
\newblock In \emph{CVPR}, 3553--3562.

\bibitem[{Huang et~al.(2022)Huang, Zhang, Heng, Shi, and Zhou}]{Huang:2022:RIF}
Huang, Z.; Zhang, T.; Heng, W.; Shi, B.; and Zhou, S. 2022.
\newblock Real-time intermediate flow estimation for video frame interpolation.
\newblock In \emph{ECCV}, 624--642.

\bibitem[{Ilg et~al.(2017)Ilg, Mayer, Saikia, Keuper, Dosovitskiy, and
  Brox}]{Ilg:2016:Flownet2}
Ilg, E.; Mayer, N.; Saikia, T.; Keuper, M.; Dosovitskiy, A.; and Brox, T. 2017.
\newblock {FlowNet} 2.0: Evolution of optical flow estimation with deep
  networks.
\newblock In \emph{CVPR}.

\bibitem[{Jain et~al.(2024)Jain, Watson, Tabellion, Ho{\l}y{\'n}ski, Poole, and
  Kontkanen}]{Jain:2024:VID}
Jain, S.; Watson, D.; Tabellion, E.; Ho{\l}y{\'n}ski, A.; Poole, B.; and
  Kontkanen, J. 2024.
\newblock Video Interpolation with Diffusion Models.
\newblock In \emph{CVPR}.

\bibitem[{Jiang et~al.(2018)Jiang, Sun, Jampani, Yang, Learned-Miller, and
  Kautz}]{Jiang:2018:SSM}
Jiang, H.; Sun, D.; Jampani, V.; Yang, M.-H.; Learned-Miller, E.; and Kautz, J.
  2018.
\newblock Super {SloMo}: {H}igh quality estimation of multiple intermediate
  frames for video interpolation.
\newblock In \emph{CVPR}, 9000--9008.

\bibitem[{Jin et~al.(2023)Jin, Wu, Chen, Chen, Koo, and Hahm}]{Jin:2023:UPR}
Jin, X.; Wu, L.; Chen, J.; Chen, Y.; Koo, J.; and Hahm, C.-h. 2023.
\newblock A unified pyramid recurrent network for video frame interpolation.
\newblock In \emph{CVPR}, 1578--1587.

\bibitem[{Kalluri et~al.(2023)Kalluri, Pathak, Chandraker, and
  Tran}]{Kalluri:2023:FLA}
Kalluri, T.; Pathak, D.; Chandraker, M.; and Tran, D. 2023.
\newblock {FLAVR}: {F}low-agnostic video representations for fast frame
  interpolation.
\newblock In \emph{WACV}, 2071--2082.

\bibitem[{Ke et~al.(2024)Ke, Obukhov, Huang, Metzger, Daudt, and
  Schindler}]{Ke:2024:RDI}
Ke, B.; Obukhov, A.; Huang, S.; Metzger, N.; Daudt, R.~C.; and Schindler, K.
  2024.
\newblock Repurposing diffusion-based image generators for monocular depth
  estimation.
\newblock In \emph{CVPR}.

\bibitem[{Kiefhaber et~al.(2024)Kiefhaber, Niklaus, Liu, and
  Schaub-Meyer}]{Kiefhaber:2024:BVF}
Kiefhaber, S.; Niklaus, S.; Liu, F.; and Schaub-Meyer, S. 2024.
\newblock Benchmarking Video Frame Interpolation.
\newblock \emph{arXiv:2403.17128 [cs.CV]}.

\bibitem[{Kim, Hwang, and Park(2024)}]{kim2024diffusehigh}
Kim, Y.; Hwang, G.; and Park, E. 2024.
\newblock Diffuse{H}igh: {T}raining-free Progressive High-Resolution Image
  Synthesis through Structure Guidance.
\newblock \emph{arXiv preprint arXiv:2406.18459}.

\bibitem[{Kingma and Ba(2014)}]{Kingma:2014:ADA}
Kingma, D.~P.; and Ba, J. 2014.
\newblock Adam: A method for stochastic optimization.
\newblock In \emph{ICLR}.

\bibitem[{Lee et~al.(2020)Lee, Kim, Chung, Pak, Ban, and Lee}]{Lee:2020:ACA}
Lee, H.; Kim, T.; Chung, T.-y.; Pak, D.; Ban, Y.; and Lee, S. 2020.
\newblock Ada{C}of: Adaptive collaboration of flows for video frame
  interpolation.
\newblock In \emph{CVPR}, 5316--5325.

\bibitem[{Li et~al.(2023)Li, Zhu, Han, Hou, Guo, and Cheng}]{Li:2023:AMT}
Li, Z.; Zhu, Z.-L.; Han, L.-H.; Hou, Q.; Guo, C.-L.; and Cheng, M.-M. 2023.
\newblock {AMT}: {A}ll-pairs multi-field transforms for efficient frame
  interpolation.
\newblock In \emph{CVPR}, 9801--9810.

\bibitem[{Liu et~al.(2024{\natexlab{a}})Liu, Zhang, Zhao, and
  Wang}]{Liu_2024_CVPR}
Liu, C.; Zhang, G.; Zhao, R.; and Wang, L. 2024{\natexlab{a}}.
\newblock Sparse Global Matching for Video Frame Interpolation with Large
  Motion.
\newblock In \emph{CVPR}, 19125--19134.

\bibitem[{Liu et~al.(2024{\natexlab{b}})Liu, Lin, Zeng, Long, Liu, Komura, and
  Wang}]{liu2023syncdreamer}
Liu, Y.; Lin, C.; Zeng, Z.; Long, X.; Liu, L.; Komura, T.; and Wang, W.
  2024{\natexlab{b}}.
\newblock Sync{D}reamer: Generating multiview-consistent images from a
  single-view image.
\newblock In \emph{ICLR}.

\bibitem[{Meyer et~al.(2018)Meyer, Djelouah, McWilliams, Sorkine-Hornung,
  Gross, and Schroers}]{Meyer:2018:PNV}
Meyer, S.; Djelouah, A.; McWilliams, B.; Sorkine-Hornung, A.; Gross, M.; and
  Schroers, C. 2018.
\newblock PhaseNet for video frame interpolation.
\newblock In \emph{CVPR}, 498--507.

\bibitem[{Meyer et~al.(2015)Meyer, Wang, Zimmer, Grosse, and
  Sorkine-Hornung}]{Meyer:2015:PBF}
Meyer, S.; Wang, O.; Zimmer, H.; Grosse, M.; and Sorkine-Hornung, A. 2015.
\newblock Phase-based frame interpolation for video.
\newblock In \emph{CVPR}, 1410--1418.

\bibitem[{Nam et~al.(2024)Nam, Lee, Kim, Kim, Cho, Kim, and Kim}]{Nam:2024:DMD}
Nam, J.; Lee, G.; Kim, S.; Kim, H.; Cho, H.; Kim, S.; and Kim, S. 2024.
\newblock Diffusion Model for Dense Matching.
\newblock In \emph{ICLR}.

\bibitem[{Niklaus and Liu(2018)}]{Niklaus:2018:CAS}
Niklaus, S.; and Liu, F. 2018.
\newblock Context-aware synthesis for video frame interpolation.
\newblock In \emph{CVPR}, 1701--1710.

\bibitem[{Niklaus and Liu(2020)}]{Niklaus:2020:SSF}
Niklaus, S.; and Liu, F. 2020.
\newblock Softmax splatting for video frame interpolation.
\newblock In \emph{CVPR}, 5437--5446.

\bibitem[{Niklaus, Mai, and Liu(2017{\natexlab{a}})}]{Niklaus:2017:VAC}
Niklaus, S.; Mai, L.; and Liu, F. 2017{\natexlab{a}}.
\newblock Video frame interpolation via adaptive convolution.
\newblock In \emph{CVPR}, 670--679.

\bibitem[{Niklaus, Mai, and Liu(2017{\natexlab{b}})}]{Niklaus:2017:VSC}
Niklaus, S.; Mai, L.; and Liu, F. 2017{\natexlab{b}}.
\newblock Video frame interpolation via adaptive separable convolution.
\newblock In \emph{ICCV}, 261--270.

\bibitem[{Niklaus, Mai, and Wang(2021)}]{Niklaus:2021:RAC}
Niklaus, S.; Mai, L.; and Wang, O. 2021.
\newblock Revisiting adaptive convolutions for video frame interpolation.
\newblock In \emph{WACV}, 1099--1109.

\bibitem[{Park, Kim, and Kim(2023)}]{Park:2023:BFL}
Park, J.; Kim, J.; and Kim, C.-S. 2023.
\newblock Bi{F}ormer: Learning bilateral motion estimation via bilateral
  transformer for {4K} video frame interpolation.
\newblock In \emph{CVPR}, 1568--1577.

\bibitem[{Park et~al.(2020)Park, Ko, Lee, and Kim}]{Park:2020:BMB}
Park, J.; Ko, K.; Lee, C.; and Kim, C.-S. 2020.
\newblock {BMBC}: Bilateral motion estimation with bilateral cost volume for
  video interpolation.
\newblock In \emph{ECCV}, 109--125.

\bibitem[{Park, Lee, and Kim(2021)}]{Park:2021:ABM}
Park, J.; Lee, C.; and Kim, C.-S. 2021.
\newblock Asymmetric bilateral motion estimation for video frame interpolation.
\newblock In \emph{ICCV}, 14539--14548.

\bibitem[{Peebles and Xie(2023)}]{Peebles:2023:SDM}
Peebles, W.; and Xie, S. 2023.
\newblock Scalable diffusion models with transformers.
\newblock In \emph{ICCV}, 4195--4205.

\bibitem[{Plack et~al.(2023)Plack, Briedis, Djelouah, Hullin, Gross, and
  Schroers}]{Plack:2023:FIT}
Plack, M.; Briedis, K.~M.; Djelouah, A.; Hullin, M.~B.; Gross, M.; and
  Schroers, C. 2023.
\newblock Frame Interpolation Transformer and Uncertainty Guidance.
\newblock In \emph{CVPR}, 9811--9821.

\bibitem[{Qian et~al.(2024)Qian, Mai, Hamdi, Ren, Siarohin, Li, Lee,
  Skorokhodov, Wonka, Tulyakov, and Ghanem}]{Qian:2024:MOT}
Qian, G.; Mai, J.; Hamdi, A.; Ren, J.; Siarohin, A.; Li, B.; Lee, H.-Y.;
  Skorokhodov, I.; Wonka, P.; Tulyakov, S.; and Ghanem, B. 2024.
\newblock Magic123: One Image to High-Quality 3{D} Object Generation Using Both
  2{D} and 3{D} Diffusion Priors.
\newblock In \emph{ICLR}.

\bibitem[{Reda et~al.(2022)Reda, Kontkanen, Tabellion, Sun, Pantofaru, and
  Curless}]{Reda:2022:FIL}
Reda, F.; Kontkanen, J.; Tabellion, E.; Sun, D.; Pantofaru, C.; and Curless, B.
  2022.
\newblock {FILM}: Frame interpolation for large motion.
\newblock In \emph{ECCV}, 250--266.

\bibitem[{Saharia et~al.(2022)Saharia, Chan, Saxena, Li, Whang, Denton,
  Ghasemipour, Gontijo~Lopes, Karagol~Ayan, Salimans, Ho, Fleet, and
  Norouzi}]{Saharia:2022:PTD}
Saharia, C.; Chan, W.; Saxena, S.; Li, L.; Whang, J.; Denton, E.~L.;
  Ghasemipour, K.; Gontijo~Lopes, R.; Karagol~Ayan, B.; Salimans, T.; Ho, J.;
  Fleet, D.~J.; and Norouzi, M. 2022.
\newblock Photorealistic text-to-image diffusion models with deep language
  understanding.
\newblock \emph{NeurIPS}, 36479--36494.

\bibitem[{Salimans and Ho(2022)}]{Salimans:2022:PDF}
Salimans, T.; and Ho, J. 2022.
\newblock Progressive Distillation for Fast Sampling of Diffusion Models.
\newblock In \emph{ICLR}.

\bibitem[{Saxena et~al.(2023)Saxena, Herrmann, Hur, Kar, Norouzi, Sun, and
  Fleet}]{Saxena:2023:TSE}
Saxena, S.; Herrmann, C.; Hur, J.; Kar, A.; Norouzi, M.; Sun, D.; and Fleet,
  D.~J. 2023.
\newblock The surprising effectiveness of diffusion models for optical flow and
  monocular depth estimation.
\newblock In \emph{NeurIPS}.

\bibitem[{Saxena et~al.(2024)Saxena, Hur, Herrmann, Sun, and
  Fleet}]{Saxena:2023:ZSM}
Saxena, S.; Hur, J.; Herrmann, C.; Sun, D.; and Fleet, D.~J. 2024.
\newblock Zero-Shot Metric Depth with a Field-of-View Conditioned Diffusion
  Model.
\newblock In \emph{ECCVW}.

\bibitem[{Shi et~al.(2024)Shi, Li, Zhang, He, Gong, and
  Zheng}]{shi2024resmaster}
Shi, S.; Li, W.; Zhang, Y.; He, J.; Gong, B.; and Zheng, Y. 2024.
\newblock Res{M}aster: Mastering High-Resolution Image Generation via
  Structural and Fine-Grained Guidance.
\newblock \emph{arXiv:2406.16476 [cs.CV]}.

\bibitem[{Shi et~al.(2022)Shi, Xu, Liu, Chen, and Yang}]{Shi:2022:VFI}
Shi, Z.; Xu, X.; Liu, X.; Chen, J.; and Yang, M.-H. 2022.
\newblock Video frame interpolation transformer.
\newblock In \emph{CVPR}, 17482--17491.

\bibitem[{Sim, Oh, and Kim(2021)}]{Sim:2021:XVF}
Sim, H.; Oh, J.; and Kim, M. 2021.
\newblock {XVFI}: extreme video frame interpolation.
\newblock In \emph{ICCV}, 14489--14498.

\bibitem[{Skorokhodov et~al.(2024)Skorokhodov, Menapace, Siarohin, and
  Tulyakov}]{skorokhodov2024hierarchical}
Skorokhodov, I.; Menapace, W.; Siarohin, A.; and Tulyakov, S. 2024.
\newblock Hierarchical Patch Diffusion Models for High-Resolution Video
  Generation.
\newblock In \emph{CVPR}, 7569--7579.

\bibitem[{Sun et~al.(2018)Sun, Yang, Liu, and Kautz}]{sun2018pwc}
Sun, D.; Yang, X.; Liu, M.-Y.; and Kautz, J. 2018.
\newblock {PWC-Net}: {CNN}s for optical flow using pyramid, warping, and cost
  volume.
\newblock In \emph{CVPR}, 8934--8943.

\bibitem[{Teed and Deng(2020)}]{teed2020raft}
Teed, Z.; and Deng, J. 2020.
\newblock {RAFT}: Recurrent all-pairs field transforms for optical flow.
\newblock In \emph{ECCV}, 402--419.

\bibitem[{Xu et~al.(2023)Xu, Liu, Vahdat, Byeon, Wang, and
  De~Mello}]{Xu:2023:OPS}
Xu, J.; Liu, S.; Vahdat, A.; Byeon, W.; Wang, X.; and De~Mello, S. 2023.
\newblock Open-vocabulary panoptic segmentation with text-to-image diffusion
  models.
\newblock In \emph{CVPR}, 2955--2966.

\bibitem[{Xue et~al.(2019)Xue, Chen, Wu, Wei, and Freeman}]{Xue:2019:VEW}
Xue, T.; Chen, B.; Wu, J.; Wei, D.; and Freeman, W.~T. 2019.
\newblock Video enhancement with task-oriented flow.
\newblock \emph{IJCV}, 127: 1106--1125.

\bibitem[{Yang, Hwang, and Ye(2023)}]{Yang:2023:ZCL}
Yang, S.; Hwang, H.; and Ye, J.~C. 2023.
\newblock Zero-shot contrastive loss for text-guided diffusion image style
  transfer.
\newblock In \emph{ICCV}, 22873--22882.

\bibitem[{Zhang et~al.(2023)Zhang, Zhu, Wang, Chen, Wu, and
  Wang}]{Zhang:2023:EMA}
Zhang, G.; Zhu, Y.; Wang, H.; Chen, Y.; Wu, G.; and Wang, L. 2023.
\newblock Extracting motion and appearance via inter-frame attention for
  efficient video frame interpolation.
\newblock In \emph{CVPR}, 5682--5692.

\bibitem[{Zheng et~al.(2024)Zheng, Guo, Deng, Han, Li, Xu, and
  Xu}]{zheng2024any}
Zheng, Q.; Guo, Y.; Deng, J.; Han, J.; Li, Y.; Xu, S.; and Xu, H. 2024.
\newblock Any-size-diffusion: Toward efficient text-driven synthesis for
  any-size {HD} images.
\newblock In \emph{AAAI}, 7571--7578.

\bibitem[{Zhou et~al.(2023)Zhou, Li, Han, and Lu}]{Zhou:2023:EMA}
Zhou, K.; Li, W.; Han, X.; and Lu, J. 2023.
\newblock Exploring motion ambiguity and alignment for high-quality video frame
  interpolation.
\newblock In \emph{CVPR}, 22169--22179.

\end{thebibliography}

\end{document}